\documentclass[journal]{IEEEtran}

\usepackage{graphicx}
\usepackage{epstopdf}
\usepackage{amsmath}
\usepackage{subcaption}

\usepackage{comment}

\usepackage{amsmath, amssymb}

\usepackage[switch]{lineno}

\begin{document}

\title{Recurrent Neural Networks  to Correct \\ Satellite Image Classification Maps}

\author{Emmanuel~Maggiori,~\IEEEmembership{Student member,~IEEE,}
        Guillaume~Charpiat, \\
        Yuliya~Tarabalka,~\IEEEmembership{Member,~IEEE,} 
        and~Pierre~Alliez

\thanks{E.~Maggiori, Y.~Tarabalka and P.~Alliez are with  Univerist\'e C\^ote d'Azur, TITANE team, Inria, 2004 Route des Lucioles, BP93 06902 Sophia Antipolis Cedex, France. E-mail: emmanuel.maggiori@inria.fr.}
\thanks{G.~Charpiat is with Tao team, Inria Saclay--\^Ile-de-France, LRI, B\^at. 660, Université Paris-Sud, 91405 Orsay Cedex, France.}

\thanks{Manuscript received ...; revised ...}}

%
%

\markboth{IEEE Transactions on Geoscience and Remote Sensing}%
{Shell \MakeLowercase{\textit{et al.}}: Bare Demo of IEEEtran.cls for IEEE Journals}
%

\maketitle

\begin{abstract}

While initially devised for image categorization, convolutional neural networks (CNNs) are being increasingly used for the pixelwise semantic labeling of images. However, the proper nature of the most common CNN architectures makes them good at recognizing but poor at localizing objects precisely. This problem is magnified in the context of aerial and satellite image labeling, where a spatially fine object outlining is of paramount importance.

Different iterative enhancement algorithms have been presented in the literature to progressively improve the coarse CNN outputs, seeking to sharpen object boundaries around real image edges. However, one must carefully design, choose and tune such algorithms. Instead, our goal is to directly learn the iterative process itself. For this, we formulate a generic iterative enhancement process inspired from partial differential equations, and observe that it can be expressed as a recurrent neural network (RNN). Consequently, we train such a network from manually labeled data for our enhancement task. In a series of experiments we show that our RNN effectively learns an iterative process that significantly improves the quality of satellite
image classification maps.

\end{abstract}

\section{Introduction}

One of the most explored problems in remote sensing is the pixelwise labeling of satellite imagery. Such a labeling is used in a wide range of practical applications, such as precision agriculture and urban planning. Recent technological developments have substantially increased the availability and resolution of satellite data. 
Besides the computational complexity issues that arise, these advances are posing new challenges in the processing of the images. Notably, the fact that large surfaces are covered introduces a significant variability in the appearance of the objects. In addition, the fine details in high-resolution images make it difficult to classify the pixels from elementary cues. For example, the different parts of an object often contrast more with each other than with other objects \cite{buildingDetection}. Using high-level contextual features thus plays a crucial role at distinguishing object classes.

Convolutional neural networks (CNNs) \cite{lecunNets} are receiving an increasing attention, due to their ability to automatically discover
relevant contextual features in image categorization problems. CNNs have already been used in the context of remote sensing \cite{mnih,paragiosBuildingDetection}, featuring powerful recognition capabilities. However, when the goal is to label images at the pixel level, the output classification maps are too coarse. For example, buildings are successfully detected but their boundaries in the classification map rarely coincide with the real object boundaries. 
We can identify two main reasons for this coarseness in the classification:

\textit{a)} There is a structural limitation of CNNs to carry out fine-grained classification. If we wish to keep a low number of learnable parameters, the ability to learn long-range contextual features comes at the cost of losing spatial accuracy, i.e., a trade-off between detection and localization. This is a well-known issue and still a scientific challenge \cite{deeplab,fcn}.

\textit{b)} In the specific context of remote sensing imagery, there is a significant lack of spatially accurate reference data for training. For example, the OpenStreetMap collaborative database provides large amounts of free-access maps over the earth, but irregular misregistrations and omissions are frequent all over the dataset. In such circumstances, CNNs cannot do better than learning rough estimates of the objects' locations, given that the boundaries are hardly located on real edges in the training set. 

Let us remark that in the particular context of high-resolution satellite imagery, the spatial precision of the classification maps is of paramount importance. Objects are small and a boundary misplaced by a few pixels significantly hampers the overall classification quality. In other application domains, such as semantic segmentation of  natural scenes, while there have been recent efforts to better shape the output objects, a high resolution output seems to be less of a priority. For example, in the popular Pascal VOC semantic segmentation dataset, there is a band of several unlabeled pixels around the objects, where accuracy is not computed to assess the performance of the  methods.

\begin{figure}
\begin{subfigure}{0.49\linewidth}
\centering
\includegraphics[scale=0.35]{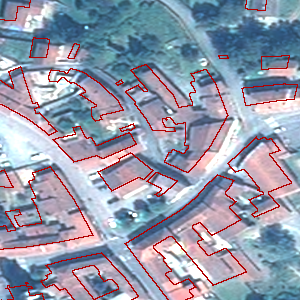}
\caption{OpenStreetMap}
\end{subfigure}
\begin{subfigure}{0.49\linewidth}
\centering
\includegraphics[scale=0.35]{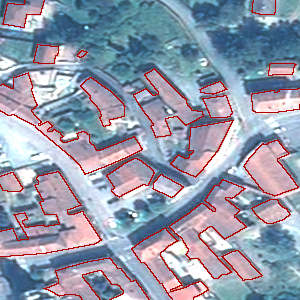}
\caption{Manual labeling}
\end{subfigure}
\caption{Samples of reference data for the \emph{building} class. Imprecise OpenStreetMap data vs manually labeled data.}
\label{f:gt_samples}
\end{figure}

There are two recent approaches to overcome the structural issues that lead to coarse classification maps. One of them is to use new types of CNN architectures, specifically designed for pixel labeling, that seek to address the detection/localization trade-off. For example, Noh et al.~\cite{deconv} duplicate a base classification CNN by attaching a reflected ``deconvolution'' network, which learns to upsample the coarse classification maps. Another tendency is to use first the base CNN as a rough classifier of the objects' locations, and then process this classification using the original image as guidance, so that the output objects better align to real image edges. For example, Zheng et al.~\cite{crfrnn} use a fully connected CRF in this manner, and Chen et al.~\cite{dt} diffuse the classification probabilities with an edge-stopping function based on image features. Both approaches have also been  adopted by the remote sensing community, mostly in the context of the ISPRS Semantic Labeling Contest, to produce fine-grained labelings of high-resolution aerial images~\cite{volpi,isprs_effective,kampf}. While all these works have certainly pushed the boundaries of CNN capabilities for pixel labeling, they assume the availability of large amounts of precisely labeled training data. This paper targets the task of dealing with more realistic datasets, seeking to provide a means to refine classification maps that are too coarse due to poor reference data.

The first scheme, i.e., the use of novel CNN architectures, seems unfeasible in the context of large-scale satellite imagery, 
due to the nature of the available training data.
 Even if an advanced architecture could eventually learn to conduct a more precise labeling, this is not useful when the training data itself is inaccurate. We thus here adopt the second strategy, reinjecting image information to an enhancement module that sharpens the coarse classification maps around the objects. To train or set the parameters of this enhancement module, as well as to validate the algorithms, we assume we can afford to manually label small amounts of data. In Fig.~\ref{f:gt_samples}(a) we show an example of imprecise data to which we have access in large quantities, and in Fig.~\ref{f:gt_samples}(b) we show a portion of manually labeled data. In our approach, the first type of data is used to train a large CNN to learn the generalities of the object classes, and the second to tune and validate the algorithm that enhances
 the coarse classification maps outputted by the CNN.

An algorithm to enhance coarse classification maps would require, on the one hand, to define 
the image features to which the objects must be attached. This is data-dependent, not every image edge being necessarily an object boundary. On the other hand, we must also decide which enhancement algorithm to use, and tune it. Besides the efforts that this requires, we could also imagine that the optimal approach would go beyond the algorithms presented in the literature. For example we could perform different types of corrections on the different classes, based on the type of errors that are often present in each of them.

Our goal is to create a system that learns the appropriate enhancement algorithm itself, instead of designing it by hand. This involves learning not only the relevant features but also the rationale behind the enhancement technique, thus intensively leveraging the power of machine learning. 

To achieve this, we first formulate a generic partial differential equation governing a broad family of iterative enhancement algorithms. This generic equation conveys the idea of progressively refining a classification map based on local cues, yet it does not provide the specifics of the algorithm. We then observe that such an equation can be expressed as a combination of common neural network layers, whose learnable parameters define the specific behavior of the algorithm. We then see the whole iterative enhancement process as a recurrent neural network (RNN). 

The RNN is provided with a small piece of manually labeled image, and trained end to end to  improve  coarse  classification maps. It automatically discovers relevant data-dependent features to enhance the classification  as well as the equations that govern every enhancement iteration.

\subsection{Related work}

A common way to tackle the aerial image labeling problem is to use classifiers such as support vector machines~\cite{svmHyperspectral} or neural networks \cite{nnHyperspectral} on the individual pixel spectral signatures (i.e., a pixel's ``color'' but not limited to RGB bands). In some cases, a few neighboring pixels are analyzed jointly to enhance the prediction and enforce the spatial smoothness of the output classification maps \cite{spectralSpatial}. Hand-designed features such as textural features have also been used \cite{nnTexture}. The use of an iterative classification enhancement process on top of hand-designed features has also been explored in the context of image labeling \cite{autocontext}.

Following the recent advent of deep learning and to address the new challenges posed by large-scale aerial imagery, Penatti et al.~\cite{doDeepFeaturesGeneralizeToRemoteSensing} used CNNs to assign aerial image patches to categories (e.g., `residential', `harbor') and Vakalopoulou et al.~\cite{paragiosBuildingDetection} addressed building detection using CNNs. Mnih \cite{mnih} and Maggiori et al.~\cite{maggiori2016fully,maggiori2017convolutional} used CNNs to learn long-range  
 contextual features
  to produce classification maps. These networks require some degree of downsampling in order to 
consider large contexts
with a reduced number of parameters. They perform well at detecting the presence of objects but do not outline them accurately.

Our work can also be related to the area of natural image semantic segmentation. Notably, fully convolutional networks (FCN) \cite{fcn}  are becoming increasingly popular to conduct pixelwise image labeling. FCN networks are made up of a stack of convolutional and pooling layers followed by so-called deconvolutional layers that upsample the resolution of the classification maps, possibly combining features at different scales. 
The output classification maps being too coarse, the authors of the Deeplab network \cite{deeplab} added a fully connected conditional random field (CRF) on top of both the FCN and the input color image, in order to enhance the classification maps.
Most of the strategies developed for natural images segmentation have been adapted to high-resolution aerial image labeling and tested on the ISPRS benchmark, including advanced FCNs~\cite{volpi} and CNNs coupled with CRFs~\cite{isprs_effective}.

Zheng et al.~\cite{crfrnn} recently reformulated the fully connected CRF of Deeplab as an RNN, and Chen et al.~\cite{dt} designed an RNN that emulates the domain transform filter \cite{gastal2011domain}. Such a  filter is used to sharpen the classification maps around image edges, which are themselves detected with a CNN. In these methods the refinement algorithm is designed beforehand and only few parameters that rule the algorithm are learned as part of the network's parameters. The innovating aspect of these approaches is that both steps (coarse classification and enhancement) can be seen as a single end-to-end network and optimized simultaneously.


Instead of predefining the algorithmic details as in previous works, we formulate a general iterative refinement algorithm through an RNN and let the network learn the specific algorithm. To our knowledge, little work has explored the idea of learning an iterative algorithm. In the context of image restoration, the preliminary work by Liu et al.~\cite{learnPDE2,learnPDE} proposed to optimize the coefficients of a linear combination of predefined terms.
Chen et al.~\cite{chen2015learning} later modeled this problem as a diffusion process and used an RNN to learn the linear filters involved as well as the coefficients of a parametrized nonlinear function. Our problem is however different, in that we use the image as guidance to update a classification map, and not to restore the image itself. Besides, while we drew inspiration on diffusion processes, we are also interested in imitating other iterative processes like active contours, thus
we do not restrict our system to 
diffusions
but consider all PDEs.

\section{Enhancing classification maps with RNNs}

\begin{figure}
\centering
\begin{subfigure}{0.32\linewidth}
\centering
\includegraphics[scale=0.37]{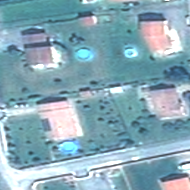}
\caption{Color image}
\end{subfigure}
\begin{subfigure}{0.32\linewidth}
\centering
\includegraphics[scale=0.37]{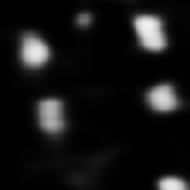}
\caption{CNN heat map}
\end{subfigure}
\begin{subfigure}{0.32\linewidth}
\centering
\includegraphics[scale=0.37]{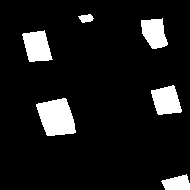}
\caption{Ground truth}
\end{subfigure}
\caption{Sample classification of buildings with a CNN. The output classification map is overly fuzzy due to the imprecision of reference data and structural limitations of CNNs.}
\label{f:heatmap}
\end{figure}

Let us assume we are given a set of score (or ``heat'') 
maps $u_k$, one for each possible class $k \in \mathcal{L}$, in a pixelwise labeling problem. 
The score of a pixel reflects the likelihood of belonging to a class, according to the classifier's predictions. The final class assigned to every pixel is the one with maximal value $u_k$. Alternatively, a softmax function can be used to interpret the results as probability scores: $P(k) =  e^{u_k} / \sum_{j \in \mathcal{L}} e^{u_j}$. Fig.~\ref{f:heatmap} shows a sample of the type of fuzzy heat map outputted by a CNN in the context of satellite image classification, for the class `building'. 

Our goal is to combine the score maps $u_k$ with information derived from the input image (e.g.,~edge features) to sharpen the scores near the real objects in order to enhance the classification.

One way to perform such a task is to progressively enhance the score maps by using  partial differential equations (PDEs). In this section we first describe different types of PDEs we could certainly imagine to design in order to solve our problem. Instead of discussing which one is the best, we then propose
 a generic iterative process
 to enhance the classification maps without specific constraints on the algorithm rationale. Finally, we show how this equation can be expressed and trained as a recurrent neural network (RNN).

\subsection{Partial differential equations (PDEs)}


We can formulate a variety of diffusion processes applied to the maps $u_k$ as partial differential equations. For example, the heat flow is described as:

\begin{equation}
\frac{\partial u_{k} (x)}{\partial t} = 
\text{div} ( \nabla u_k (x) ),
\label{eq:heatFlow}
\end{equation}
where $\text{div}(\cdot)$ denotes the divergence operator in the spatial domain of $x$. Applying such a diffusion process in our context would smooth out the heat maps. 
Instead, our goal is to design an image-dependent smoothing process that aligns the heat maps to the image features. A natural way of doing this is to modulate the gradient in Eq.~\ref{eq:heatFlow} by a scalar function $g(x,I)$ that depends on the input image $I$:
\begin{equation}
\frac{\partial u_{k} (x)}{\partial t} =  \text{div} ( g(I,x) \nabla u_k (x) ).
\label{eq:peronaMalik}
\end{equation}
Eq.~\ref{eq:peronaMalik} is similar to the Perona-Malik diffusion \cite{peronaMalik} with the exception that Perona-Malik uses the smoothed function itself to guide the diffusion. $g(I,x)$ denotes an edge-stopping function that takes low values  near borders of $I(x)$ in order to slow down the smoothing process there. 

Another possibility would be to consider a more general variant in which $g(I,x)$ is replaced by a matrix $D(I,x)$, acting as a diffusion tensor that redirects the  flow based on image properties instead of just slowing it down near edges:
\begin{equation}
\frac{\partial u_{k} (x)}{\partial t} = \text{div} ( D(I,x) \nabla u_k (x) ).
\label{eq:anisDiff}
\end{equation}
This formulation relates to the so-called anisotropic diffusion process \cite{weickertBook}.

Alternatively, one can draw 
inspiration from the level set framework. For example, the geodesic active contours technique formulated as level sets translates into: 
\begin{equation}
\frac{\partial u_{k} (x)}{\partial t} = |\nabla u_{k}(x)| \text{div} \left( g(I,x) \frac{\nabla u_k (x)}{|\nabla u_k(x)|} \right).
\label{eq:gac}
\end{equation}
Such 
a 
formulation favors the zero level set to align with minima of $g(I,x)$ \cite{gac}. 
Schemes based on Eq.~\ref{eq:gac} could then be used to improve heat maps $u_k$, provided they are scaled 
so that segmentation boundaries match zero levels.

As shown above, many different PDE approaches can be devised to enhance classification maps. However, several choices must be made to select the appropriate PDE and tailor it to our problem. 

 For example, one must choose the edge-stopping function $g(I,x)$ in Eqs.~\ref{eq:peronaMalik}, \ref{eq:gac}. Common choices are exponential or rational functions on the image gradient~\cite{peronaMalik}, which in turn requires to set an edge-sensitivity parameter. Extensions to the original Perona-Malik approach could also be considered, such as a popular regularized variant that computes the gradient on a Gaussian-smoothed version of the input image \cite{weickertBook}. 
In the case of opting for anisotropic diffusion, one must design $D(I,x)$.

 Instead of using trial and error to perform such design, our goal is to let a machine learning approach discover
by
 itself a useful iterative process for our task.

\subsection{A generic classification enhancement process}

PDEs are usually discretized in space by using finite differences, which represent derivatives as discrete convolution filters. We build upon this scheme to write a generic discrete formulation of an enhancement iterative process.

Let us consider that we take as
input a score map $u_k$ (for class $k$) and, in the most general case, an arbitrary number of feature maps $\{g_1,...,g_p\}$ derived from image~$I$. 
In order to perform differential operations, of the type $\{\frac{\partial}{\partial x},\frac{\partial}{\partial y}, \frac{\partial^2}{\partial x \partial y}, \frac{\partial^2}{\partial x^2},...\}$, we consider convolution kernels $\{M_1,M_2,...\}$ and $\{N_1^j, N_2^j,...\}$ to be applied to the heat map $u_k$ and to the features $g_j$ derived from image $I$, respectively. While we could certainly directly provide a bank of filters $M_i$ and $N_i^j$ in the form of Sobel operators, Laplacian operators, etc., we may simply let the system learn the required filters. 
 We group all the feature maps that result from applying these convolutions, in a single set: 
\begin{equation}
\Phi(u_k, I) = \left\{ M_i \ast u_k,\; \;N_l^j \ast g_j(I)\; ;\;\; \forall i,j,l\right\}.
\label{eq:functionals}
\end{equation}
Let us now define a generic discretized scheme as:
\begin{equation}
\frac{\partial u_{k} (x)}{\partial t} =  
 f_k  \big( \; \Phi(u_k,I)(x)\; \big),
\label{eq:spaceDiscretization_new}
\end{equation}
where $f_k$ is a function that takes as input the values of all the features in $\Phi(u_k,I)$ at an image point $x$, and combines them.
While convolutions $M_i$ and $N_i^j$  convey the ``spatial'' reasoning, e.g., gradients, $f_k$ captures the combination of these elements, such as the products in Eqs.~\ref{eq:peronaMalik} and \ref{eq:gac}.
 

Instead of deriving an arbitrary number of possibly complex features $N_i^j \ast g_j(I)$ from image $I$, we can think of a simplified scheme in which we directly operate on $I$, by considering only convolutions: $N_i \ast I$. The  list of functionals considered in Eq.~\ref{eq:spaceDiscretization_new} is then 
\begin{equation}
\Phi(u_k, I) = \Big\{ M_i \ast u_k, \;\;N_j \ast I \; ; \;\; \forall i,j \Big\}
\label{eq:functionals_simplified}
\end{equation}
and consists only of convolutional kernels directly applied to the heat maps $u_k$ and to the image $I$.
From now on, we here stick to this simpler formulation, yet we acknowledge that it might be eventually useful to work on a higher-level representation rather than on the input image itself. 
Note that if one restricts functions $f_k$ in Eq.~\ref{eq:spaceDiscretization_new} to be linear, we still obtain the set of all linear PDEs. We consider \emph{any} function $f_k$, introducing non-linearities.

PDEs are usually discretized in time, taking the form:
\begin{equation}
u_{k,t+1} (x) = u_{k,t} (x) 
 + \delta u_{k,t} (x),
\label{eq:timeDiscretization}
\end{equation}
where $\delta u_{k,t}
$ denotes 
the overall update of $u_{k,t}$ at time~$t$.

Note that  the convolution filters in Eqs.~\ref{eq:functionals} and \ref{eq:functionals_simplified} are 
class-agnostic:
$M_i$, $N_j$ and $N_l^j$ do not depend on $k$,
while $f_k$ may be a different function for each class~$k$. Function $f_k$ thus determines the contribution of each feature to the equation, contemplating the case in which a different evolution might be optimal for each of the classes, even if just in terms of
a time-step factor.
In the next section we detail a way to learn the update functions $\delta u_{k,t}$
from training data.

\subsection{Iterative processes as RNNs}
\label{s:PDEasRNN}

\begin{figure*}
\centering
\includegraphics[scale=0.67]{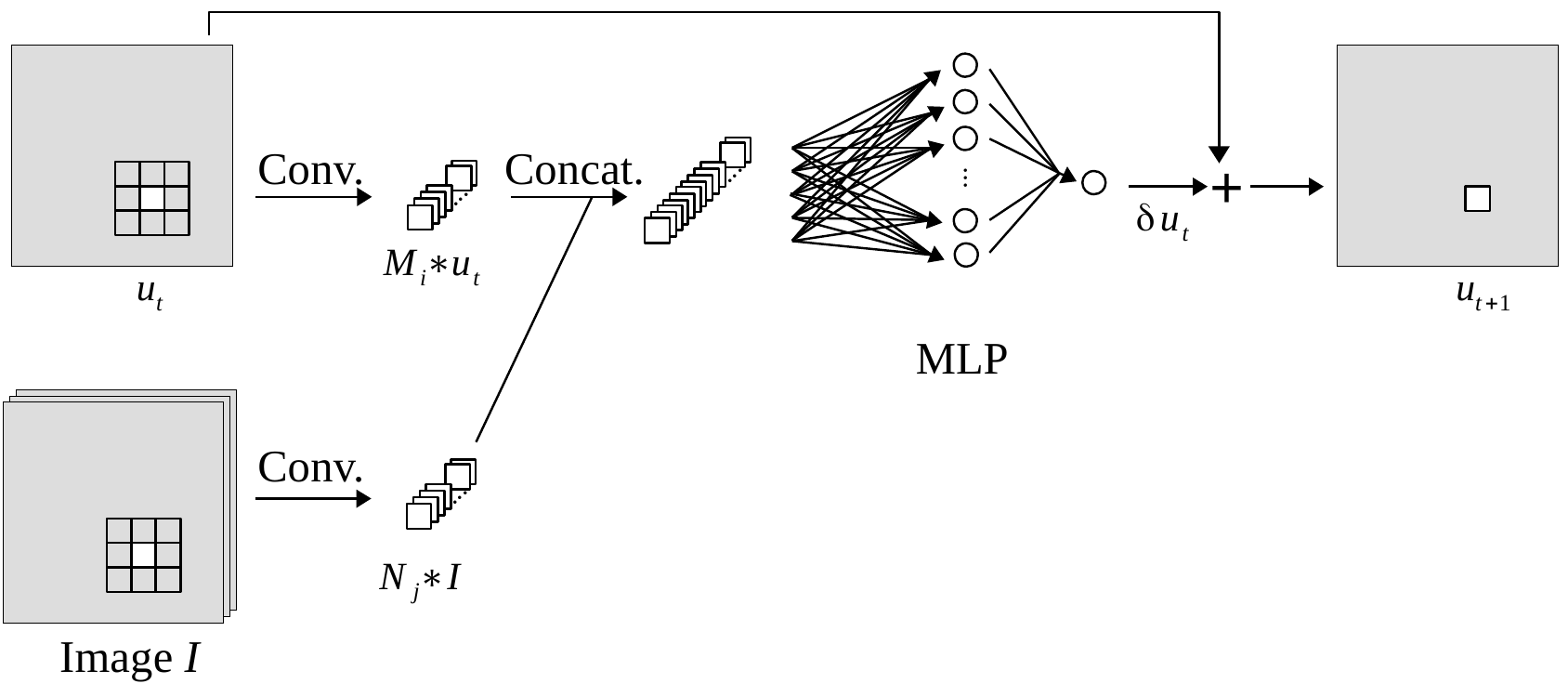}

\caption{One enhancement iteration represented as common neural network layers. Features are extracted both from the input image $I$ and the heat map of the previous iteration $u_t$. These are then concatenated and inputted to an MLP, which computes the update $\delta u_t$. The heat map $u_t$ is added to the update $\delta u_t$ to yield the modified map $u_{t+1}$.}
\label{f:rnn_oneiter}
\end{figure*}

We now show that the generic iterative process can be implemented as an RNN, and thus trained from labeled data. This stage requires to provide the system with a piece of accurately labeled ground truth (see e.g., Fig.~\ref{f:gt_samples}b).

Let us first show that one 
 iteration, as defined in Eqs.~\ref{eq:spaceDiscretization_new}-\ref{eq:timeDiscretization}, can be expressed in terms of common neural network layers.  
   Let us focus on a single pixel for a specific class, simplifying the notation from $u_{k,t} (x)$ to $u_t$. Fig.~\ref{f:rnn_oneiter} illustrates the proposed network architecture. 
Each iteration takes as input the image $I$ and a given heat map $u_t$ to enhance at time $t$.
In the first iteration, $u_t$ is the initial coarse heat map to be improved, outputted by another pre-trained neural network in our case. From the heat map $u_t$ we derive a series of filter responses, which correspond to $M_i \ast u_t$ in Eq.~\ref{eq:functionals_simplified}. These responses are found by computing the dot product between a set of filters $M_i$ and the values of $u_{k,t}(\cdot)$ in a spatial neighborhood of a given point. Analogously, a set of filter responses are computed at the same spatial location on the input image, corresponding to the different $N_j \ast I$ of Eq.~\ref{eq:functionals_simplified}. 
These operations are convolutions when performed densely in space, $N_j\ast I$ and $M_i\ast u_t$ being feature maps of the filter responses.

These filters are then ``concatenated'', forming a pool of 
features $\Phi$ coming from both the input image and the heat map, as in Eq.~\ref{eq:functionals_simplified}, and inputted to $f_k$ in Eq.~\ref{eq:spaceDiscretization_new}. 
We must now learn the function $\delta u_t$ that describes how the heat map $u_t$ is updated at iteration $t$ (cf.~Eq.~\ref{eq:timeDiscretization}), based on these features.

Eq.~\ref{eq:spaceDiscretization_new} does not introduce specifics about function $f_k$.
 In (\ref{eq:heatFlow})-(\ref{eq:gac}), for example, it includes products between different terms, but we could certainly imagine other functions. 
  We therefore model $\delta u_t$ through a multilayer perceptron (MLP), because it can approximate any function within a bounded error. We include one hidden layer with nonlinear activation functions followed by an output neuron with a linear activation (a typical configuration for regression problems), although other MLP architectures could be used. 
Applying this MLP densely is equivalent to performing convolutions with $1\times1$ kernels at every layer. The implementation to densely label entire images is then straightforward.

The value of $\delta u_t$ is then added to $u_{t}$ in order to generate the updated map $u_{t+1}$. This addition is performed pixel by pixel in the case of a dense input. Note that although we could have removed this addition and let the MLP directly output the updated map $u_{t+1}$, we opted for this architecture since it is more closely related to the equations and better conveys the intention of a progressive refinement of the classification map. Moreover, learning $\delta u_t$ instead of $u_{t+1}$ has a significant advantage at training time: a random initialization of the networks' parameters centered around zero means that the initial RNN represents an iterative process close to the identity (with some noise). Training uses the asymmetry induced by this noise to progressively move from the identity to a more useful iterative process.

\begin{figure*}
\centering
\includegraphics[scale=0.85]{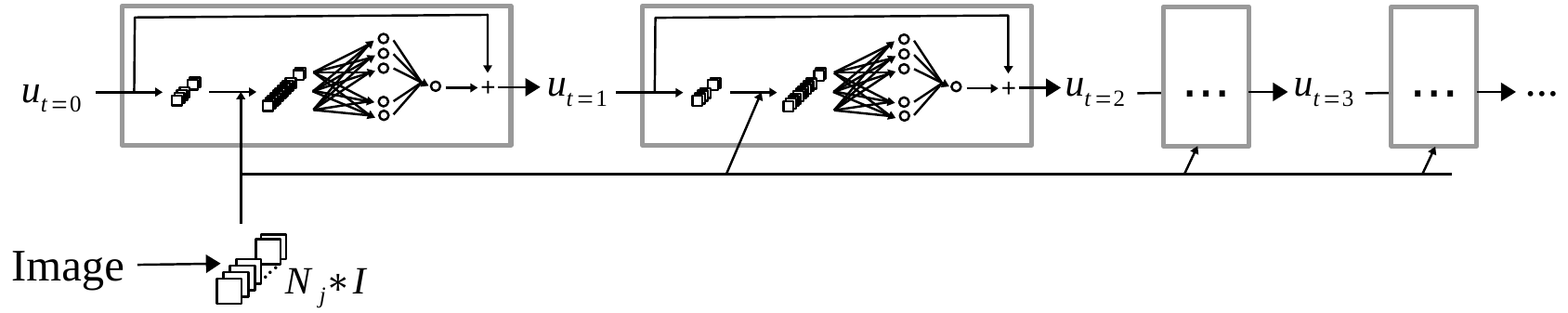}
\caption{The modules of Fig.~\ref{f:rnn_oneiter} are stacked (while sharing parameters) to implement the iterative process as an RNN.}
\label{f:rnn_manyiter}
\end{figure*}

The overall iterative process is implemented by unrolling a finite number of iterations, as illustrated in Fig.~\ref{f:rnn_manyiter}, under the constraint that the parameters are shared among all iterations. Such sharing is enforced at training time by a simple modification to the back-propagation training algorithm where the derivatives of every instance of a weight at different iterations are averaged \cite{backpropThroughTime}. Note that issues with vanishing or exploding gradients may arise when too many iterations are unrolled, an issue inherent to deep network architectures. 
Note also that the spatial features are shared across the classes, while a different MLP is learned for each of them, following Eq.~\ref{eq:spaceDiscretization_new}. As depicted by Fig.~\ref{f:rnn_manyiter} and conveyed in the equations, the features extracted from the input image are independent of the iteration.

The RNN of Fig.~\ref{f:rnn_manyiter} represents then a dynamical system that iteratively improves the class heat maps. Training such an RNN amounts to finding the optimal dynamical system for our enhancement task.

\section{Implementation details}
\label{s:implDetails}

We first describe the CNN used to produce the coarse predictions, then detail our RNN. The
  network architecture was implemented using
 Caffe 
deep learning
 library \cite{caffe}.

Our coarse prediction CNN is based on a previous remote sensing network presented by Mnih \cite{mnih}. 
We create a fully convolutional~\cite{fcn} version of Mnih's network, since recent remote sensing work has shown the theoretical and practical advantages of this type of architecture~\cite{kampffmeyer2016semantic,maggiori2016fully}.
The CNN takes 3-band color image patches at 1m$^2$ resolution and produces as many heat maps as classes considered. The resulting four-layer architecture is as follows: 64 conv.~filters ($12\times 12$, stride 4) $\rightarrow$ 128 conv.~filters ($3\times 3$) $\rightarrow$ 128 conv.~filters ($3\times 3$) $\rightarrow$ 3 conv.~filters ($9\times9$). Since the first convolution is performed with a stride of 4, the resulting feature maps have a quarter of the input resolution. Therefore, a deconvolutional layer \cite{fcn} is added on top to upsample the classification maps to the original resolution. The activation functions used in the hidden layers are rectified linear units.
This network is trained on patches randomly selected from the training dataset. We group 64 patches with classification maps of size $64\times 64$ into mini-batches (following \cite{mnih}) to estimate the gradient of the network's parameters and back-propagate them. Our loss function is the cross-entropy between the target and predicted class probabilities. Stochastic gradient descent is used for optimization, with learning rate 0.01, momentum 0.9 and an L2 weight regularization of 0.0002. We did not however optimized these parameters nor the networks' architectures.

We now detail the implementation of the RNN described in Sec.~\ref{s:PDEasRNN}. Let us remark that at this stage we fix the weights of the initial coarse CNN and the manually labeled tile is used to train the RNN only.  Our RNN learns 32 $M_i$ and 32 $N_j$ filters, both of spatial dimensions $5 \times 5$.
An independent MLP is learned for every class, using 32 hidden neurons each and with rectified linear activations, while $M_i$ and $N_j$ filters are shared across the different classes (in accordance to Equations~\ref{eq:spaceDiscretization_new} and~\ref{eq:functionals_simplified}). This highlights the fact that $M_i$ and $N_j$ capture low-level features while the MLPs convey class-specific behavior. We unroll five RNN iterations, which enables us to significantly improve the classification maps without exhausting our GPU's memory. 
Training is performed on random patches and with the cross-entropy loss function, as done with the coarse CNN. The employed gradient descent algorithm is AdaGrad~\cite{adagrad}, which exhibits a faster convergence in our case, using a base learning rate of 0.01 (higher values make the loss diverge). All weights are initialized randomly by sampling from a distribution that depends on the number of neuron inputs~\cite{xavier}.
 We trained the RNN for 50,000 iterations, until observing convergence of the training loss, which took around four hours on a single GPU.

\section{Experiments}

We perform our experiments on images acquired by a Pl\'eiades satellite over the area of Forez, France. An RGB color image is used, obtained by pansharpening \cite{pansharpening} the satellite data, which provides a spatial resolution of 0.5 $\text{m}^2$. Since the networks described in Sec.~\ref{s:implDetails} are specifically designed for 1 $\text{m}^2$ resolution images, we downsample the Pl\'eiades images before feeding them to our networks and bilinearly upsample the outputs.

\begin{figure}
\centering
\begin{subfigure}{0.5\linewidth}
\centering
\includegraphics[scale=0.04]{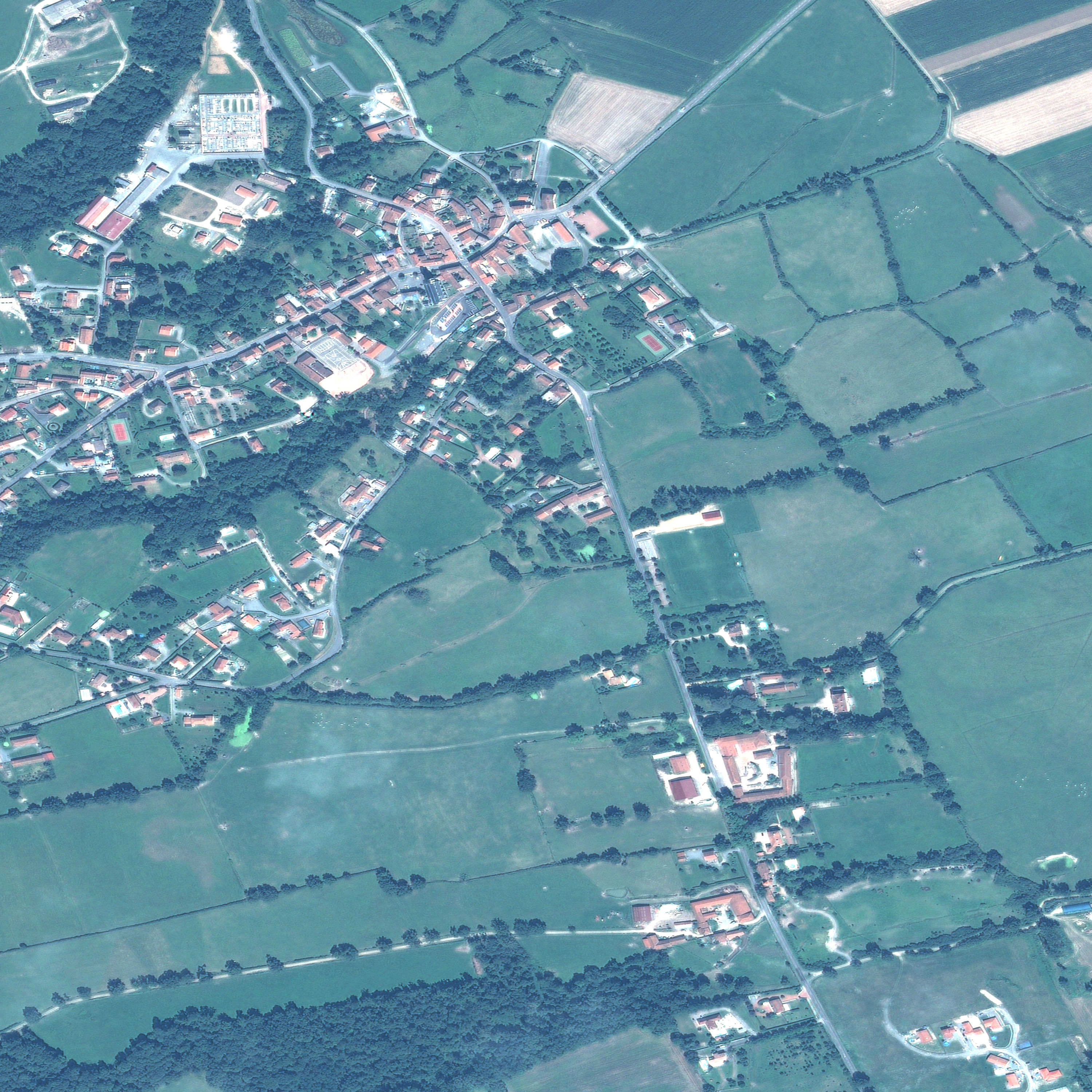}
\end{subfigure}%
\begin{subfigure}{0.5\linewidth}
\centering
\includegraphics[scale=0.04]{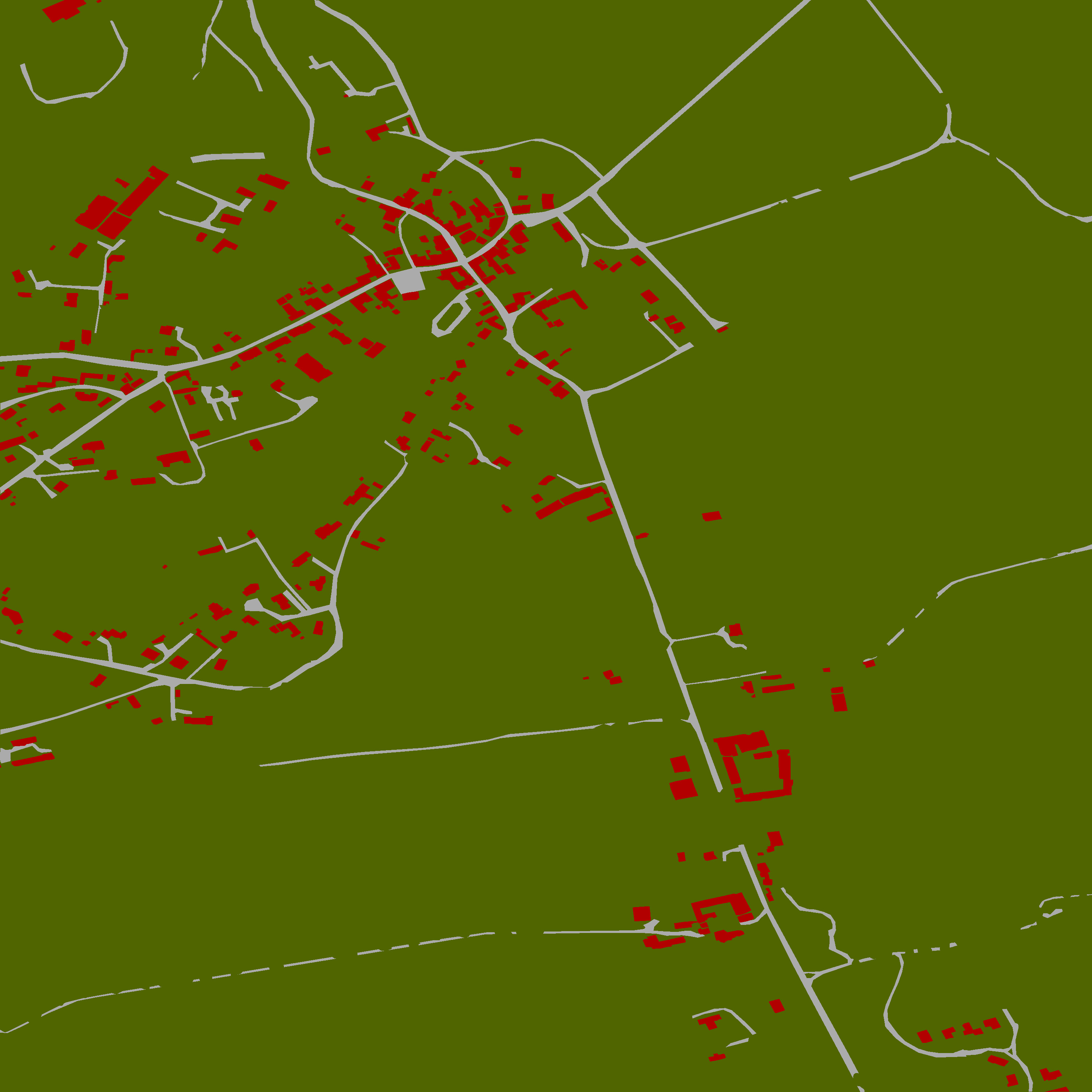}
\end{subfigure}%
\caption{Manually labeled tile used to train the RNN for the classification enhancement task.}
\label{f:fineTuningTile}
\end{figure}

\begin{figure*}
\centering
\begin{subfigure}[t]{0.2\linewidth}
\centering
\begin{subfigure}{0.49\linewidth}
\centering

\includegraphics[scale=0.231]{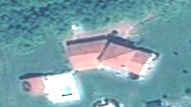}

\vspace{1.5pt}
\includegraphics[scale=0.202]{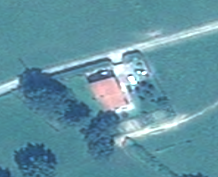}

\vspace{1.5pt}
\includegraphics[scale=0.317]{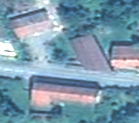}

\vspace{1.5pt}
\includegraphics[scale=0.44]{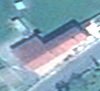}

\vspace{1.5pt}
\includegraphics[scale=0.427]{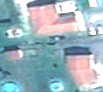}

\vspace{1.5pt}
\includegraphics[scale=0.254]{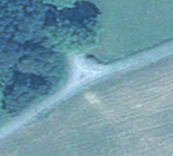}

\caption*{Color}
\end{subfigure}
\begin{subfigure}{0.48\linewidth}
\centering

\includegraphics[scale=0.231]{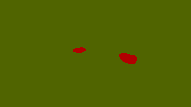}

\vspace{1.5pt}
\includegraphics[scale=0.202]{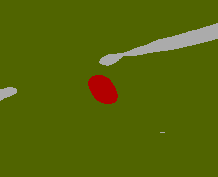}

\vspace{1.5pt}
\includegraphics[scale=0.317]{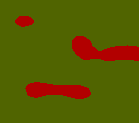}

\vspace{1.5pt}
\includegraphics[scale=0.44]{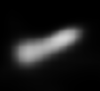}

\vspace{1.5pt}
\includegraphics[scale=0.427]{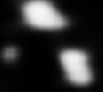}

\vspace{1.5pt}
\includegraphics[scale=0.254]{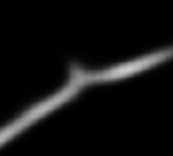}

\caption*{CNN map}
\end{subfigure}
(RNN input)
\end{subfigure}
\hspace{10pt}
\begin{subfigure}{0.41\linewidth}
\centering
\begin{subfigure}{0.235\textwidth}
\centering
\includegraphics[scale=0.231]{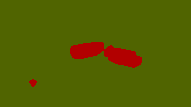}

\vspace{1.5pt}
\includegraphics[scale=0.202]{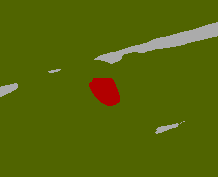}

\vspace{1.5pt}
\includegraphics[scale=0.317]{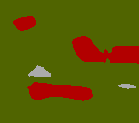}

\vspace{1.5pt}
\includegraphics[scale=0.44]{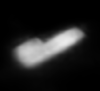}

\vspace{1.5pt}
\includegraphics[scale=0.427]{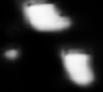}

\vspace{1.5pt}
\includegraphics[scale=0.254]{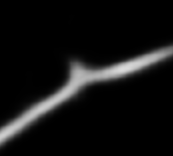}
\end{subfigure}
\begin{subfigure}{0.235\textwidth}
\centering
\includegraphics[scale=0.231]{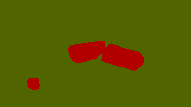}

\vspace{1.5pt}
\includegraphics[scale=0.202]{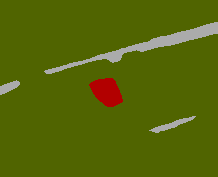}

\vspace{1.5pt}
\includegraphics[scale=0.317]{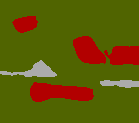}

\vspace{1.5pt}
\includegraphics[scale=0.44]{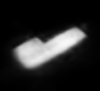}

\vspace{1.5pt}
\includegraphics[scale=0.427]{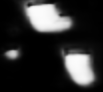}

\vspace{1.5pt}
\includegraphics[scale=0.254]{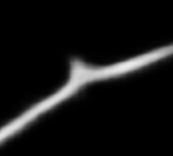}
\end{subfigure}
\begin{subfigure}{0.235\textwidth}
\centering
\includegraphics[scale=0.231]{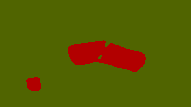}

\vspace{1.5pt}
\includegraphics[scale=0.202]{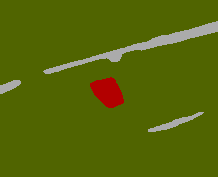}

\vspace{1.5pt}
\includegraphics[scale=0.317]{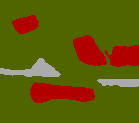}

\vspace{1.5pt}
\includegraphics[scale=0.44]{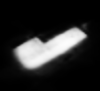}

\vspace{1.5pt}
\includegraphics[scale=0.427]{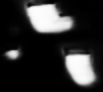}

\vspace{1.5pt}
\includegraphics[scale=0.254]{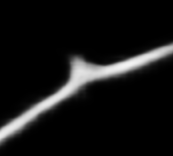}
\end{subfigure}
\begin{subfigure}{0.235\textwidth}
\centering
\includegraphics[scale=0.231]{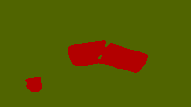}

\vspace{1.5pt}
\includegraphics[scale=0.202]{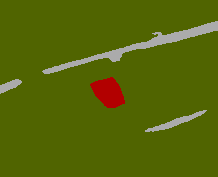}

\vspace{1.5pt}
\includegraphics[scale=0.317]{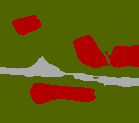}

\vspace{1.5pt}
\includegraphics[scale=0.44]{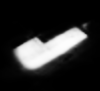}

\vspace{1.5pt}
\includegraphics[scale=0.427]{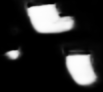}

\vspace{1.5pt}
\includegraphics[scale=0.254]{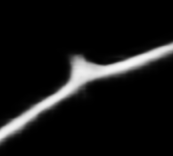}
\end{subfigure}

\caption*{--- Intermediate RNN iterations ---}
\end{subfigure}
\begin{subfigure}{0.12\linewidth}
\centering
\includegraphics[scale=0.231]{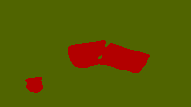}

\vspace{1.5pt}
\includegraphics[scale=0.202]{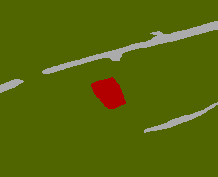}

\vspace{1.5pt}
\includegraphics[scale=0.317]{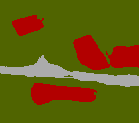}

\vspace{1.5pt}
\includegraphics[scale=0.44]{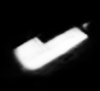}

\vspace{1.5pt}
\includegraphics[scale=0.427]{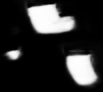}

\vspace{1.5pt}
\includegraphics[scale=0.254]{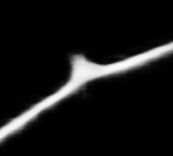}

\caption*{RNN output}
\end{subfigure}
\begin{subfigure}{0.12\linewidth}
\centering
\includegraphics[scale=0.231]{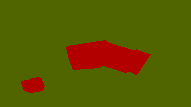}

\vspace{1.5pt}
\includegraphics[scale=0.202]{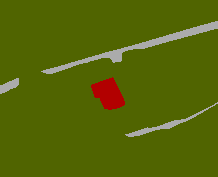}

\vspace{1.5pt}
\includegraphics[scale=0.317]{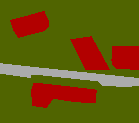}

\vspace{1.5pt}
\includegraphics[scale=0.44]{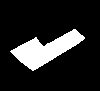}

\vspace{1.5pt}
\includegraphics[scale=0.427]{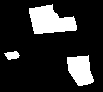}

\vspace{1.5pt}
\includegraphics[scale=0.254]{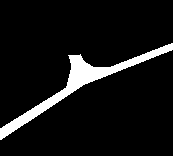}

\caption*{\mbox{Ground truth}}
\end{subfigure}

\caption{Evolution of fragments of classification maps (top rows) and single-class fuzzy scores (bottom rows) through RNN iterations. The classification maps are progressively sharpened around object's edges.}
\label{f:frags_iterations}

\end{figure*}


From this image we selected an area with OpenStreetMap (OSM) \cite{osm} coverage to create a 22.5 $\text{km}^2$ training dataset for the classes \emph{building}, \emph{road} and \emph{background}. The reference data was obtained by rasterizing the raw OSM maps.  
Misregistrations and omissions are present all over the dataset (see, e.g., Fig.~\ref{f:gt_samples}a). Buildings tend to be misaligned or omitted, while many roads in the ground truth are not visible in the image (or the other way around). Moreover, since OSM's roads are represented by polylines, we set a fixed road width of 7 m to rasterize this class (following~\cite{mnih}), which makes their classification particularly challenging.  This dataset is used to train the initial coarse CNNs.

\begin{figure*}
\centering
\begin{subfigure}{0.32\linewidth}
\centering
\includegraphics[scale=0.055]{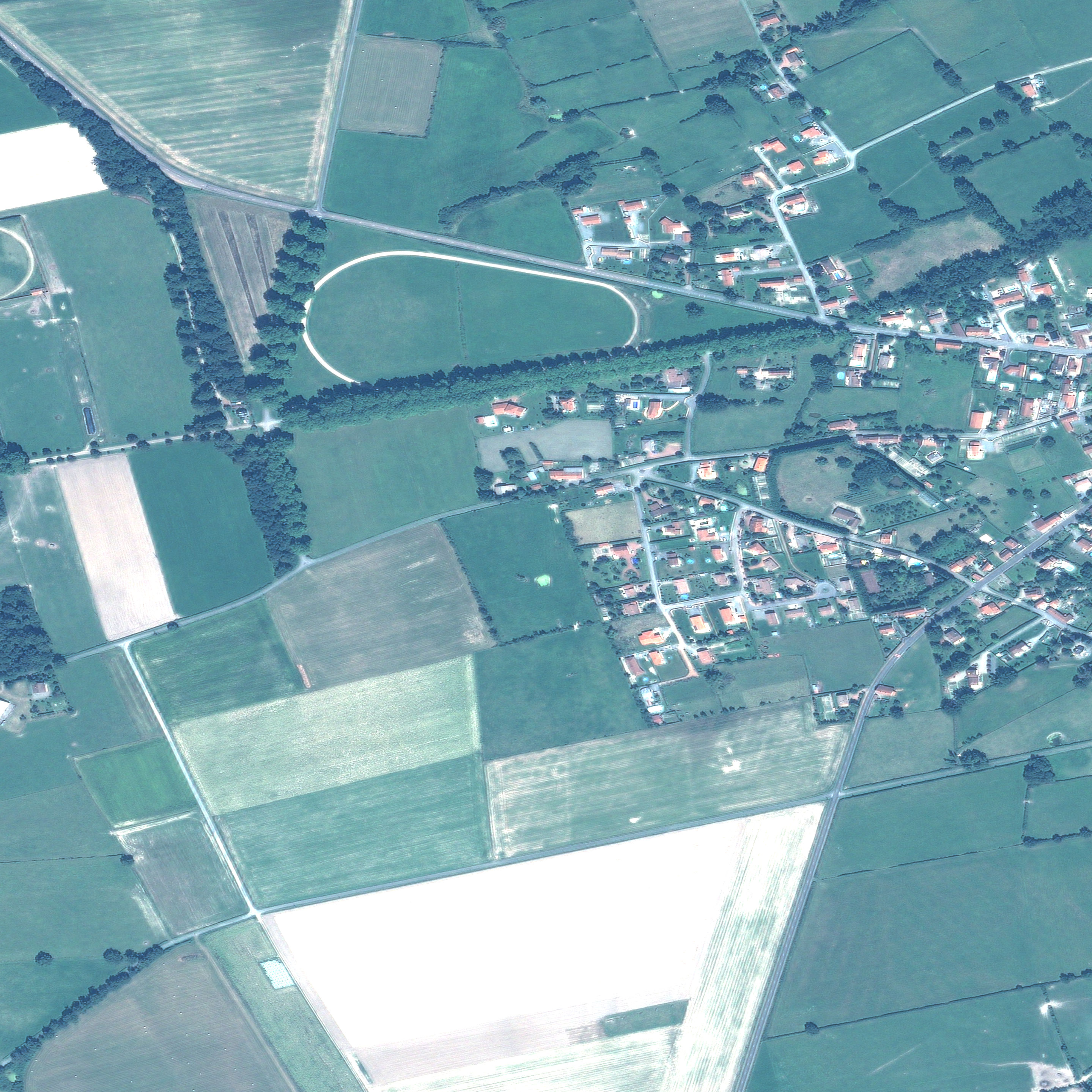}
\caption{Color image}
\end{subfigure}
\begin{subfigure}{0.32\linewidth}
\centering
\includegraphics[scale=0.055]{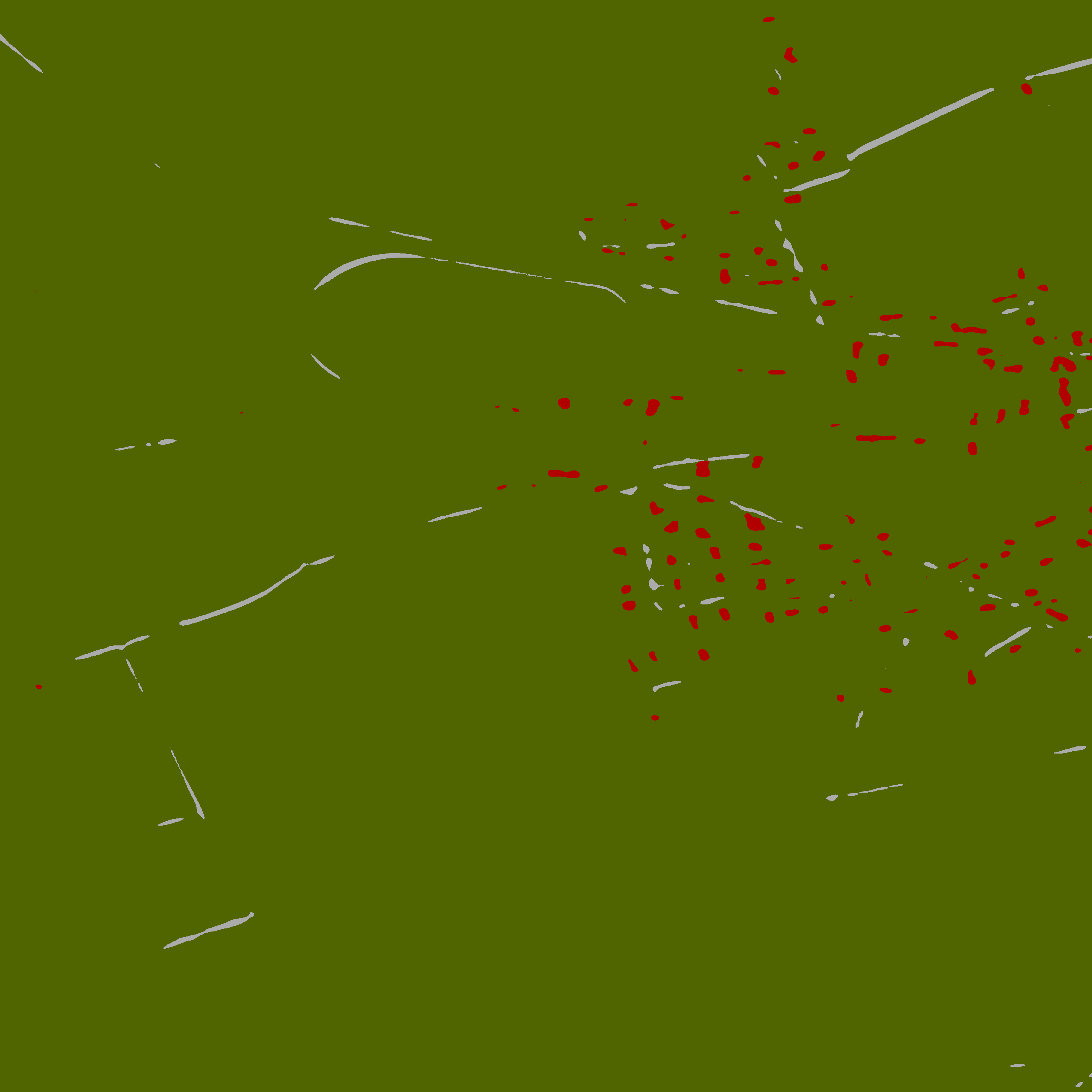}
\caption{Coarse CNN}
\end{subfigure}
\begin{subfigure}{0.32\linewidth}
\centering
\includegraphics[scale=0.055]{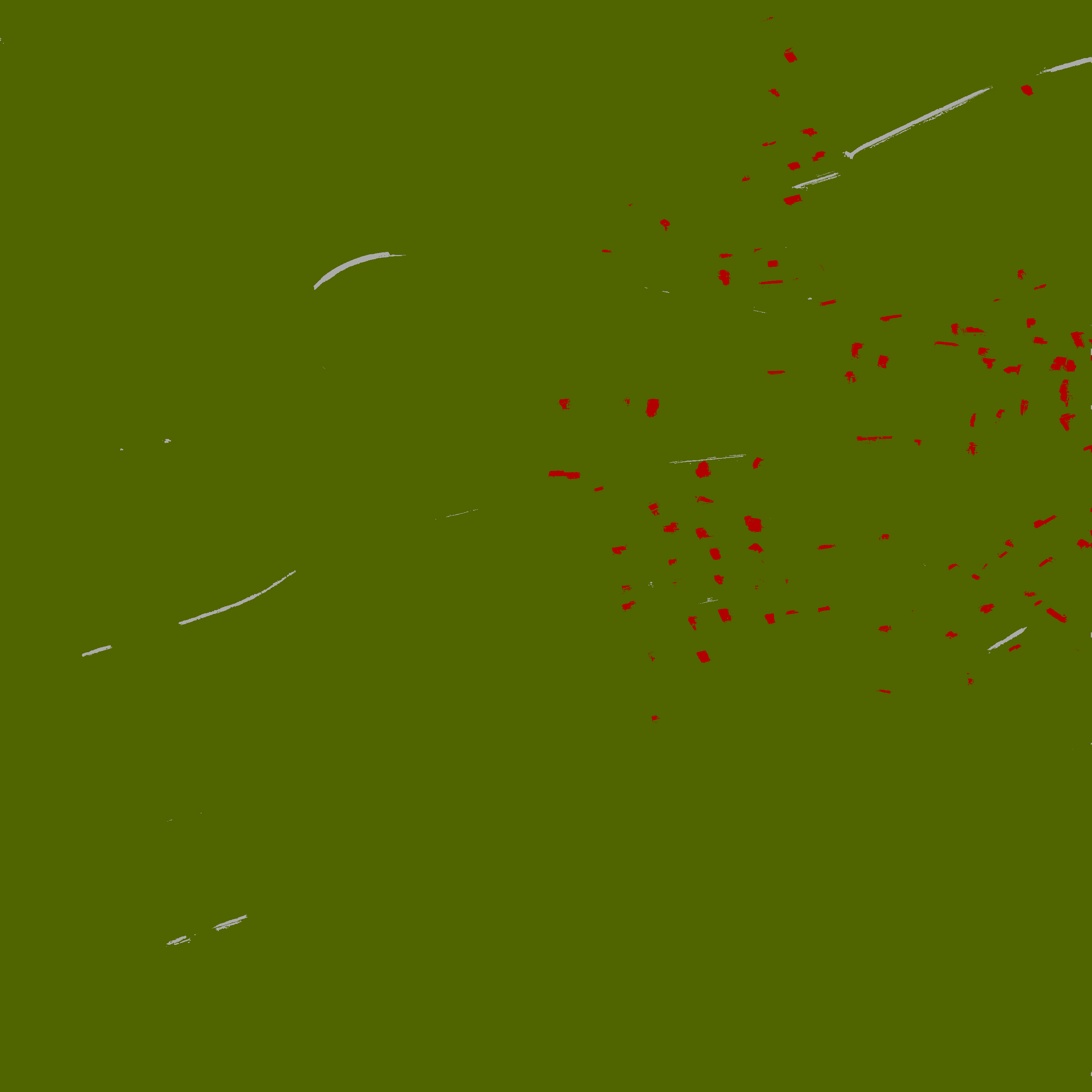}
\caption{CNN+CRF}
\end{subfigure}
\begin{subfigure}{0.32\linewidth}
\centering
\includegraphics[scale=0.055]{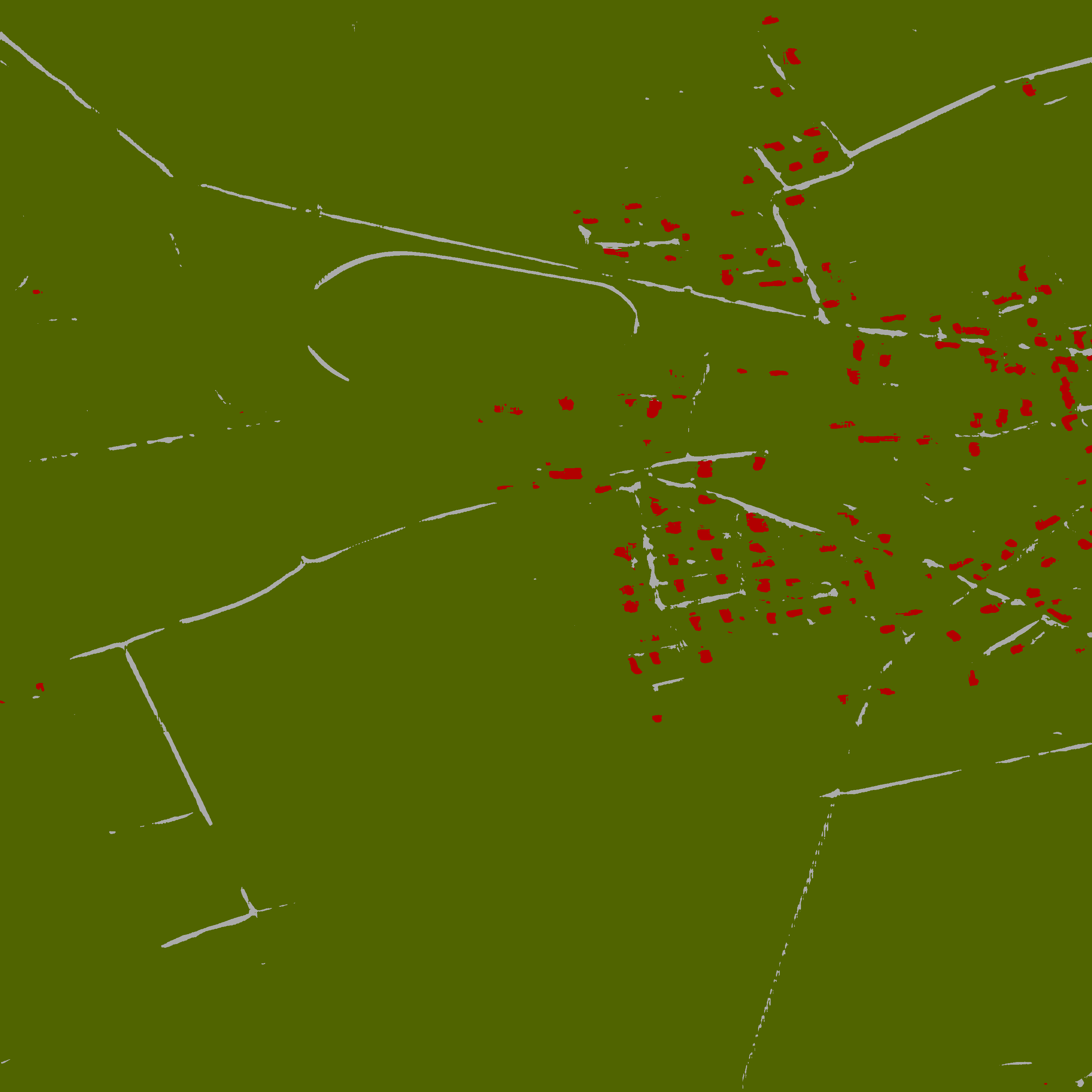}
\caption{Class-agnostic CNN+RNN}
\end{subfigure}
\begin{subfigure}{0.32\linewidth}
\centering
\includegraphics[scale=0.055]{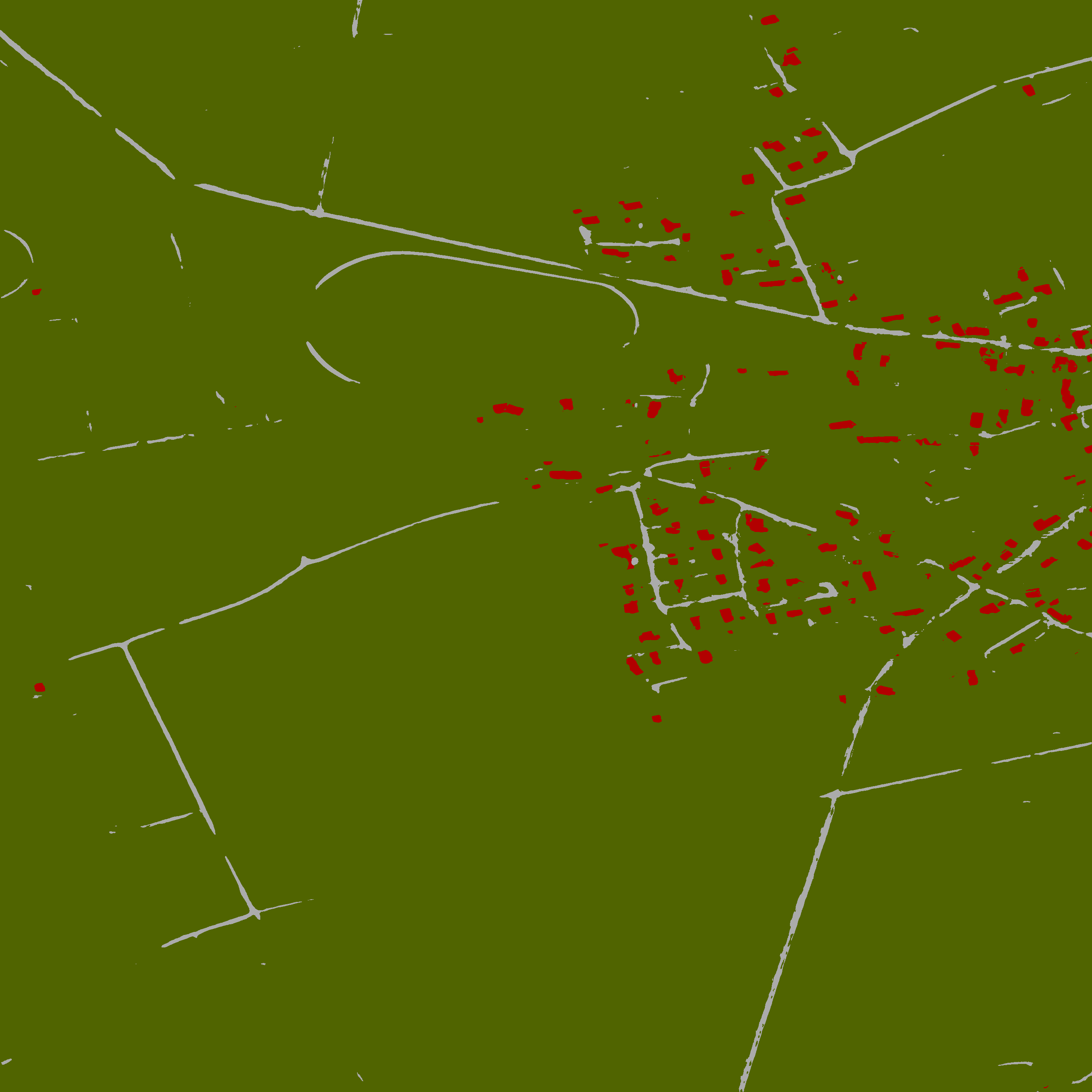}
\caption{CNN+RNN}
\end{subfigure}
\begin{subfigure}{0.32\linewidth}
\centering
\includegraphics[scale=0.055]{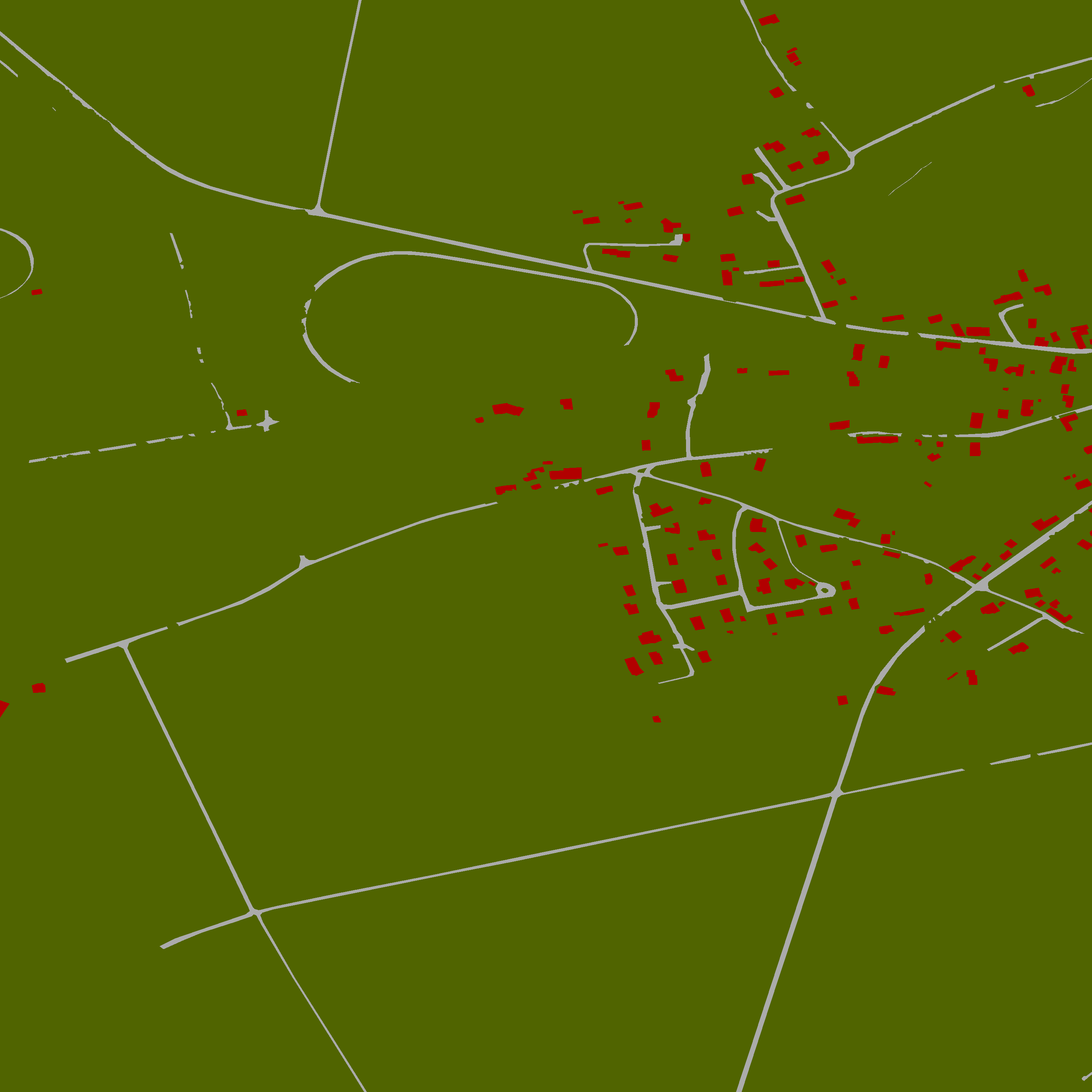}
\caption{Ground truth}
\end{subfigure}

\caption{Visual comparison on a Pl\'eiades satellite image tile of size 3000$\times$3000 covering 2.25 km$^2$.}
\label{f:large_results}

\end{figure*}

We manually labeled two 2.25 $\text{km}^2$ tiles to train and test the RNN at enhancing the predictions of the coarse network. We denote them by enhancement and test sets, respectively. Note that our RNN system must discover an algorithm to refine an existing classification map, and not to conduct the classification itself, hence a smaller training set should be sufficient for this stage. The enhancement set is depicted in Fig.~\ref{f:fineTuningTile} while the test set is shown in Figs.~\ref{f:large_results}(a)/(f).

\begin{figure}
\begin{subfigure}{\linewidth}
\centering
\setlength{\tabcolsep}{4pt}
\small \begin{tabular}{ccc|ccc}
 & Overall & Mean  &  \multicolumn{3}{c}{Class-specific IoU} \\ 
Method & \hspace{-5pt} accuracy \hspace{-5pt} &  IoU & Build. & Road & Backg. \\ 
\hline 
CNN & 96.72 & 48.32 & 38.92 & 9.34 & 96.69 \\ 
CNN+CRF & 96.96 & 44.15 & 29.05 & 6.62 & 96.78 \\  
CNN+RNN$^=$ & 97.78 & 65.30 & 59.12 & 39.03 & 97.74  \\ 
CNN+RNN & \textbf{98.24} & \textbf{72.90} & \textbf{69.16} & \textbf{51.32} & \textbf{98.20} \\ 
\hline 
\end{tabular} 
\caption{Numerical comparison (in \%)}
\end{subfigure}

\begin{subfigure}{\linewidth}
\centering
\includegraphics[scale=0.55]{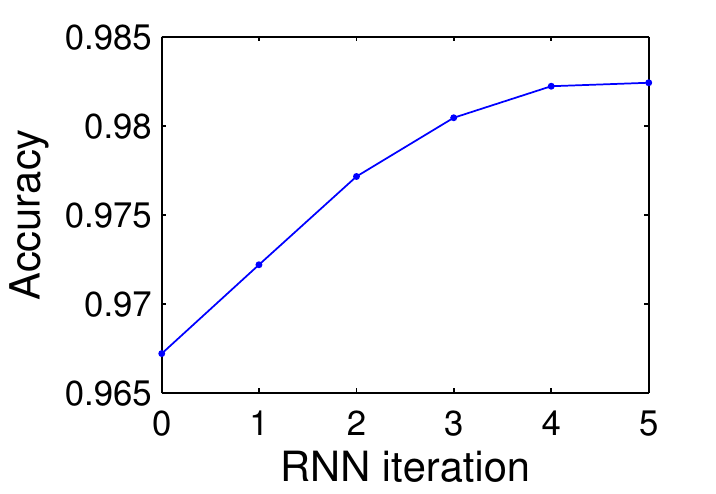}
\includegraphics[scale=0.55]{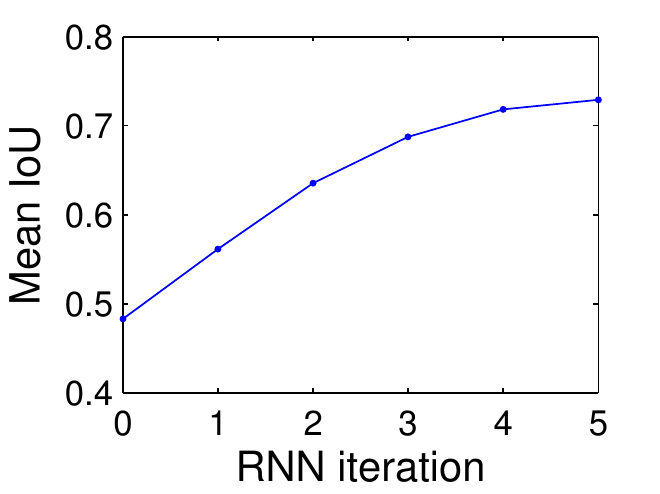}
\caption{Evolution through RNN iterations}
\end{subfigure}

\caption{Quantitative evaluation on Pl\'eaiades images test set over Forez, France.}
\vspace{10pt}
\label{f:quantitative}
\end{figure}

In the following, we report the results obtained by using the proposed method on the \mbox{Pl\'eiades} dataset.
Fig.~\ref{f:frags_iterations} provides closeups of results on different fragments of the test dataset. The initial and final maps (before and after the RNN enhancement) are depicted, as well as the intermediate results through the RNN iterations. We show both a set of final classification maps and some single-class fuzzy probability maps. We can observe that as the RNN iterations go by, the classification maps are refined and the objects better align to image edges. The fuzzy probabilities become more confident, sharpening object boundaries. 
To quantitatively assess this improvement we compute two measures on the test set: the overall accuracy (proportion of correctly classified pixels) and the  intersection over union (IoU)~\cite{fcn}. Mean IoU has become the standard in semantic segmentation since it is more reliable in the presence of imbalanced classes (such as \emph{background} class, which is included to compute the mean)~\cite{acc}. As summarized in the table of Fig.~\ref{f:quantitative}(a), the performance of the original coarse CNN (denoted by \textsc{cnn}) is significantly improved by attaching our RNN (\textsc{cnn+rnn}). Both measures increase monotonously along the intermediate RNN iterations, as depicted in Fig.~\ref{f:quantitative}(b).

\begin{figure}
\begin{subfigure}{0.25\linewidth}
\centering
\includegraphics[scale=0.114]{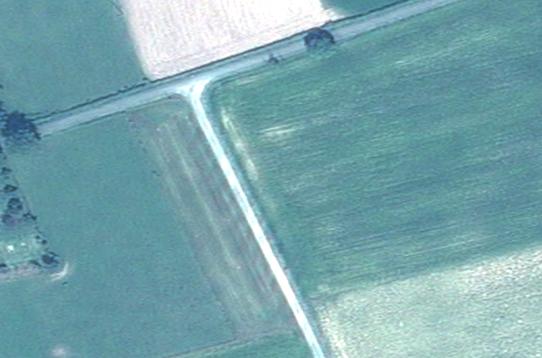}

\vspace{1.5pt}
\includegraphics[scale=0.183]{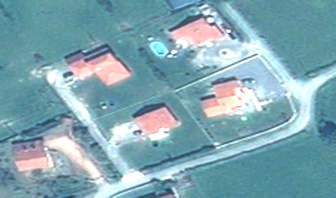}

\vspace{1.5pt}
\includegraphics[scale=0.102]{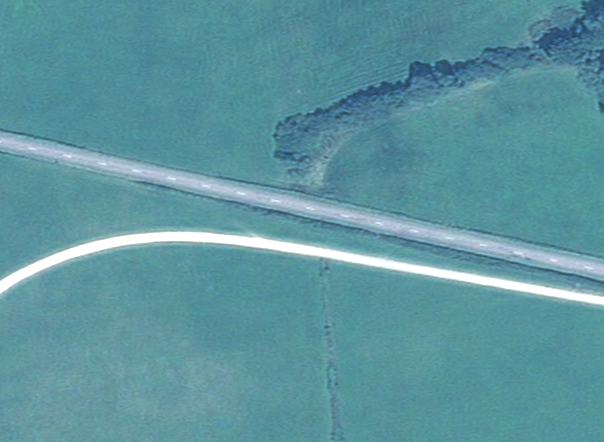}

\caption*{Color image}
\end{subfigure}%
\begin{subfigure}{0.25\linewidth}
\centering
\includegraphics[scale=0.114]{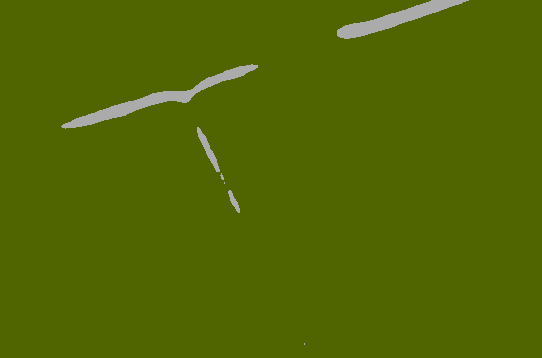}

\vspace{1.5pt}
\includegraphics[scale=0.183]{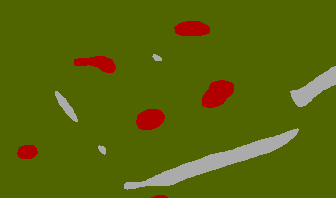}

\vspace{1.5pt}
\includegraphics[scale=0.102]{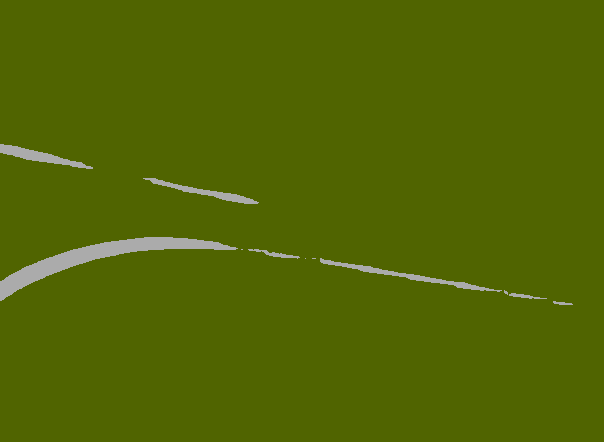}

\caption*{Coarse CNN }
\end{subfigure}%
\begin{subfigure}{0.25\linewidth}
\centering
\includegraphics[scale=0.114]{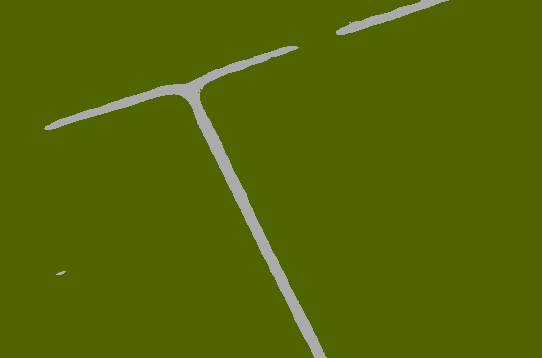}

\vspace{1.5pt}
\includegraphics[scale=0.183]{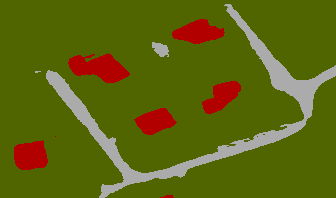}

\vspace{1.5pt}
\includegraphics[scale=0.102]{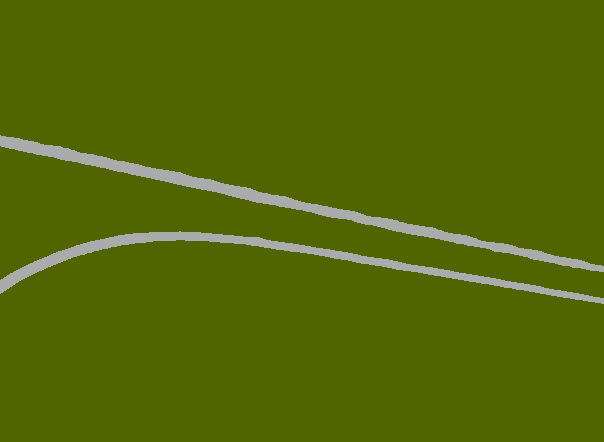}

\caption*{RNN output}
\end{subfigure}%
\begin{subfigure}{0.25\linewidth}
\centering
\includegraphics[scale=0.114]{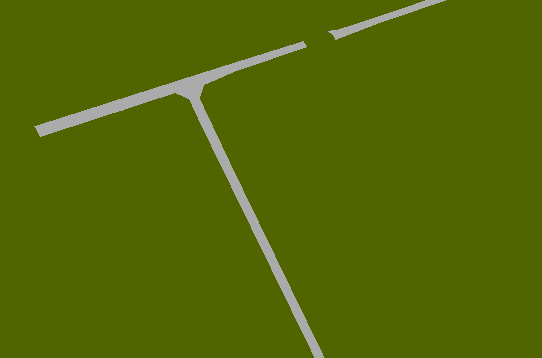}

\vspace{1.5pt}
\includegraphics[scale=0.183]{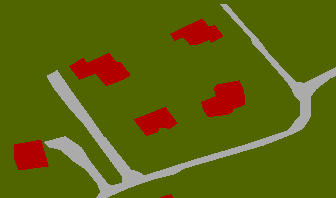}

\vspace{1.5pt}
\includegraphics[scale=0.102]{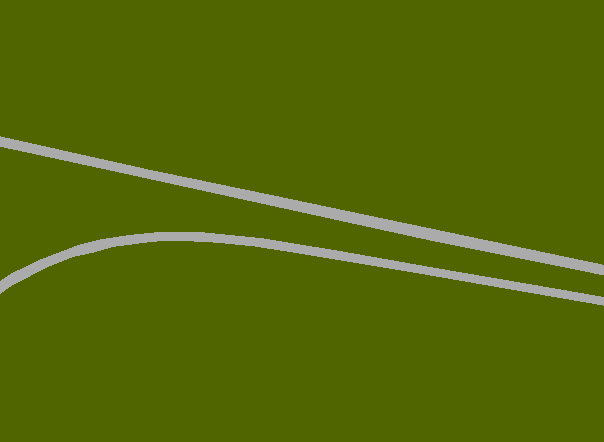}

\caption*{Ground truth}
\end{subfigure}%

\caption{Initial coarse classifications and the enhanced maps by using RNNs.}
\label{f:beforeAfter}
\end{figure}

The initial classification of roads has an overlap of less than 10\% with the roads in the ground truth, as shown by its individual IoU. The RNN makes them emerge from the background class, now overlapping the ground truth roads by over 50\%. Buildings also become better aligned to the real boundaries, going from less than 40\%  to over 70\% overlap with the ground truth buildings.
This constitutes a multiplication of the IoU by a factor of 5 for roads and 2 for buildings, which indicates a significant improvement at outlining and not just detecting objects.

Additional visual fragments before and after the RNN refinement are shown in Fig.~\ref{f:beforeAfter}. We can observe in the last row how the iterative process learned by the RNN both thickens and narrows the roads depending on the location.

\begin{figure*}
\captionsetup[subfigure]{justification=centering}
\centering
\begin{subfigure}[t]{0.161\linewidth}
\centering
\includegraphics[scale=0.22]{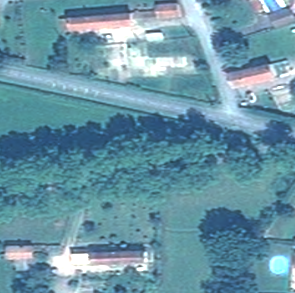}

\vspace{1.5pt}
\includegraphics[scale=0.405]{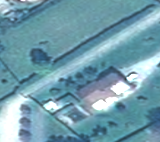}

\vspace{1.5pt}
\includegraphics[scale=0.156]{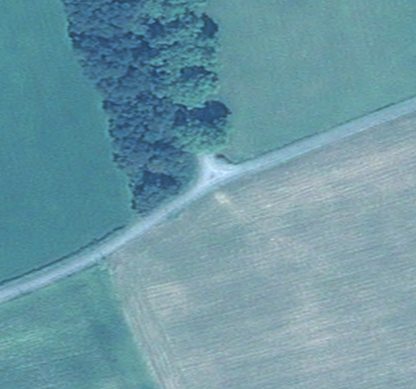}

\caption*{Color image}
\end{subfigure}
\begin{subfigure}[t]{0.161\linewidth}
\centering
\includegraphics[scale=0.22]{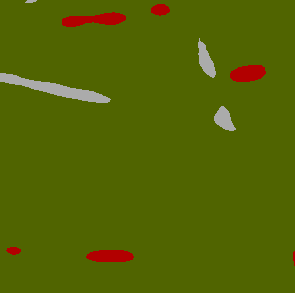}

\vspace{1.5pt}
\includegraphics[scale=0.405]{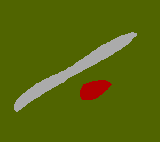}

\vspace{1.5pt}
\includegraphics[scale=0.156]{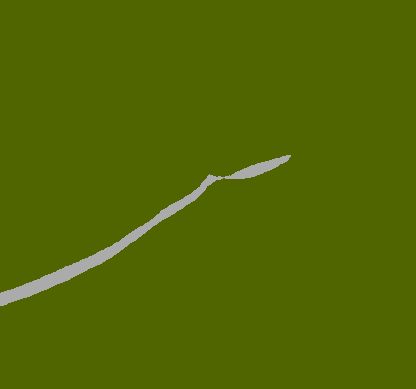}

\caption*{Coarse CNN}
\end{subfigure}
\begin{subfigure}[t]{0.161\linewidth}
\centering
\includegraphics[scale=0.22]{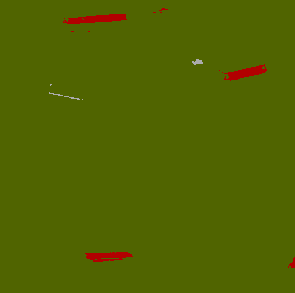}

\vspace{1.5pt}
\includegraphics[scale=0.405]{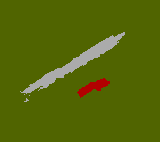}

\vspace{1.5pt}
\includegraphics[scale=0.156]{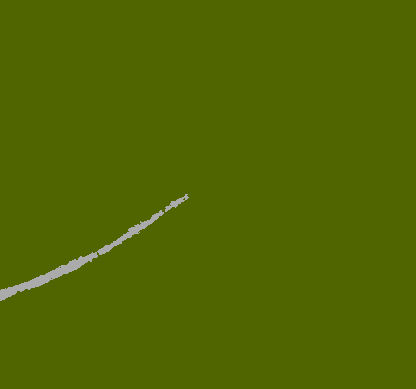}

\caption*{CNN+CRF}
\end{subfigure}
\begin{subfigure}[t]{0.161\linewidth}
\centering
\includegraphics[scale=0.22]{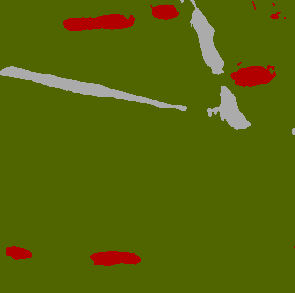}

\vspace{1.5pt}
\includegraphics[scale=0.405]{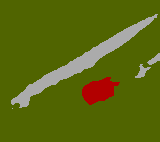}

\vspace{1.5pt}
\includegraphics[scale=0.156]{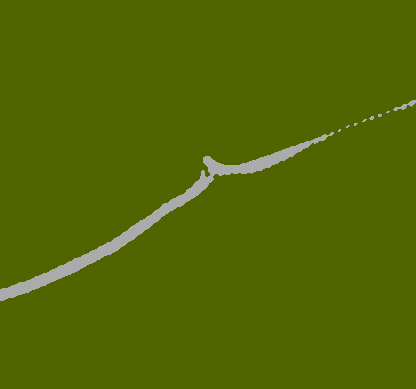}

\caption*{Class-agnostic CNN+RNN}
\end{subfigure}
\begin{subfigure}[t]{0.161\linewidth}
\centering
\includegraphics[scale=0.22]{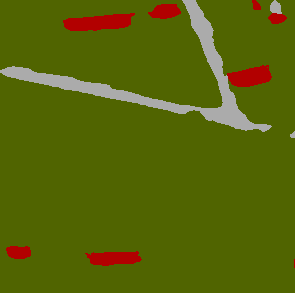}

\vspace{1.5pt}
\includegraphics[scale=0.405]{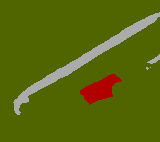}

\vspace{1.5pt}
\includegraphics[scale=0.156]{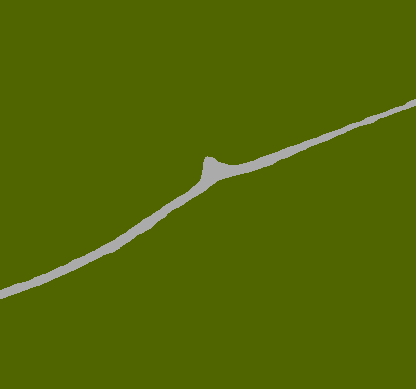}

\caption*{CNN+RNN}
\end{subfigure}
\begin{subfigure}[t]{0.161\linewidth}
\centering
\includegraphics[scale=0.22]{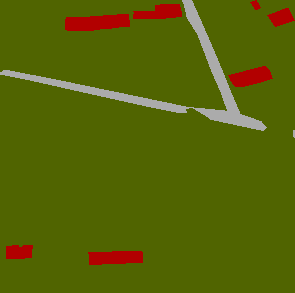}

\vspace{1.5pt}
\includegraphics[scale=0.405]{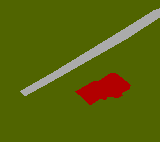}

\vspace{1.5pt}
\includegraphics[scale=0.156]{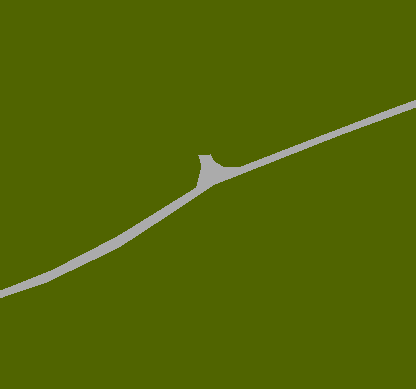}

\caption*{Ground truth}
\end{subfigure}

\caption{Visual comparison on closeups of the Pl\'eiades dataset.}
\label{f:comparison}
\end{figure*}

We also compare our RNN to the approach
  in~\cite{deeplab} (here denoted by \textsc{cnn+crf}), where a fully-connected CRF is coupled both to the input image and the coarse CNN output, in order to refine the predictions. 
This is the idea behind the so-called Deeplab network, which constitutes one of the most important current baselines in the semantic segmentation community.
 While the CRF itself could also be implemented as an RNN~\cite{crfrnn}, we here stick to the original 
 formulation because the CRF as RNN idea is only interesting if we want to train the system end to end (i.e., together with the coarse prediction network). In our case we wish to leave the coarse network as is, 
 otherwise we risk overfitting it to this much smaller set. 
 We thus simply use the CRF as in~\cite{deeplab} and tune the energy parameters by performing a grid search using the enhancement set as a reference. Five  iterations of inference on the fully-connected CRF were preformed in every case.
 
To further analyze our method, we also consider an alternative enhancement RNN in which the weights of the MLP are shared across the different classes (which we denote by ``class-agnostic \textsc{cnn+rnn}''). This forces the system to learn the same function to update all the classes, instead of a class-specific function.

 Numerical results are included in the table of Fig.~\ref{f:quantitative}(a) and the classification maps are shown in in Fig.~\ref{f:large_results}. Close-ups of these maps are included in Fig.~\ref{f:comparison} to facilitate comparison. The CNN+CRF approach does sharpen the maps but this often occurs around the wrong edges. It also makes small objects disappear in favor of larger objects (usually the background class) when edges are not well marked, which explains the mild increase in overall accuracy but the decrease in mean IoU. While the class-agnostic CNN+RNN outperforms the CRF, both quantitative and visual results are beaten by the CNN+RNN, supporting the importance of learning a class-specific enhancement function. 

To validate the importance of using a recurrent architecture, and following Zheng et al.~\cite{crfrnn}, we retrained our system considering every iteration of the RNN as an independent step with its own parameters. After training for the same number of iterations, it yields a lower performance on the test set compared to the RNN and a higher performance on the training set. If we keep on training, the non-recurrent network still enhances its training accuracy while performing poorly on the test set, implying a significant degree of overfitting with this variant of the
 architecture. This provides evidence that constraining our network to learn an iterative enhancement process is crucial for its success.

\subsection{Feature visualization}

Though it is difficult to interpret the overall function learned by the RNN, especially the part of the multi-layer perceptron, there are some things we expect to find if we analyze the spatial filters $M_i$ and $N_j$ learned by the RNN (see Eq.~\ref{eq:functionals_simplified}). Carrying out this analysis is a way of validating the behavior of the network.

The iterative process learned by the RNN should combine information from both the heat maps and the image at every iteration (since the heat maps constitute the prior on where the objects are located, and the image guides the enhancement of these heat maps). 
A logical way of enhancing the classification is to align the high-gradient areas of the heat maps with the object boundaries. We expect then to find derivative operators among the filters $N_j$ applied to the heat maps. Concerning  the image filters $N_j$, we expect to find data-dependent filters (e.g.,~image edge detectors) that help identify the location of object boundaries.
 
\begin{figure}
\begin{subfigure}{\linewidth}
\begin{subfigure}[t]{0.32\linewidth}
\centering
Heat maps

\includegraphics[scale=0.17]{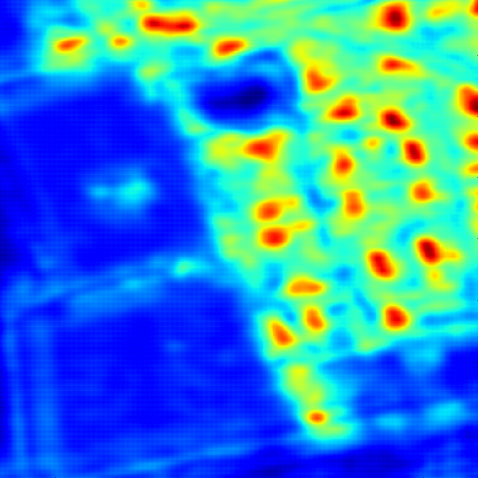}

\vspace{1.5pt}
\includegraphics[scale=0.17]{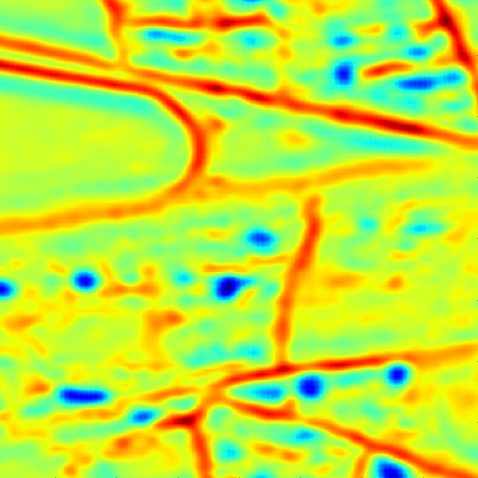}
\end{subfigure}
\begin{subfigure}[t]{0.65\linewidth}
\centering
Feature responses

\begin{subfigure}{0.49\linewidth}
\centering
\includegraphics[scale=0.17]{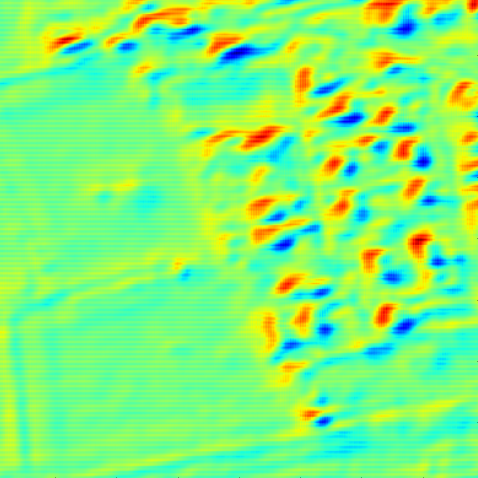}

\vspace{1.5pt}
\includegraphics[scale=0.17]{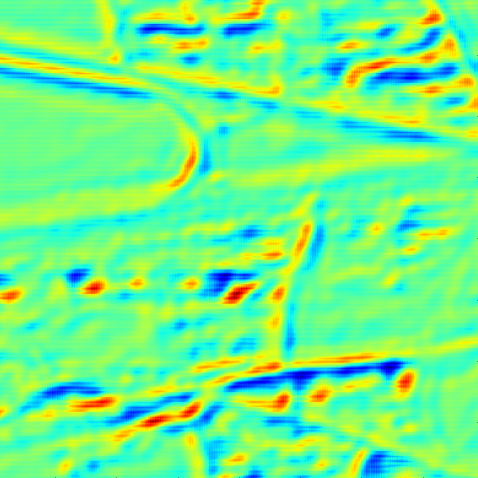}

{\small $M_1$}
\end{subfigure}
\begin{subfigure}{0.49\linewidth}
\centering
\includegraphics[scale=0.17]{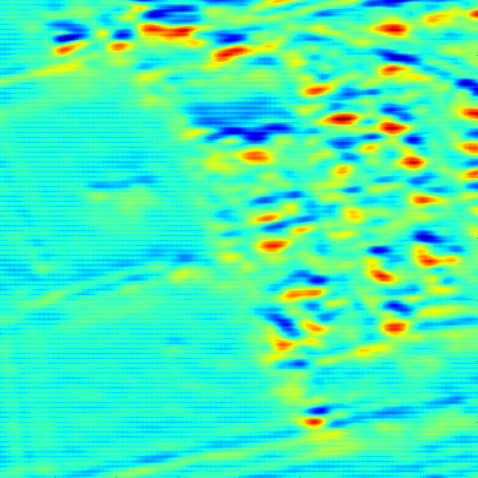}

\vspace{1.5pt}
\includegraphics[scale=0.17]{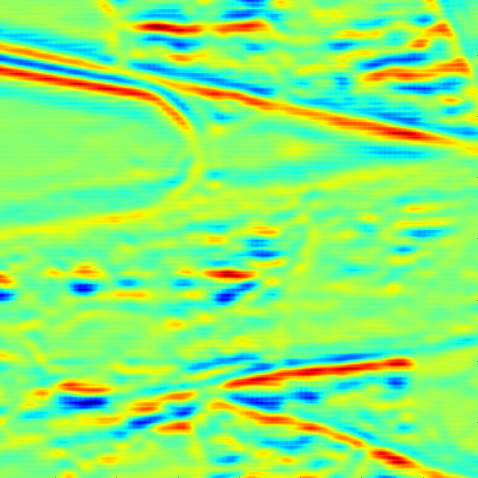}
{\small $M_2$}
\end{subfigure}
\end{subfigure}
\caption{Filter $M_1$ acts like a gradient operator in the South-East direction and $M_2$ in the North direction (top: building, bottom: road).}
\end{subfigure}

\vspace{3pt}
\begin{subfigure}{\linewidth}
\begin{subfigure}[t]{0.32\linewidth}
\centering
Color image

\includegraphics[scale=0.17]{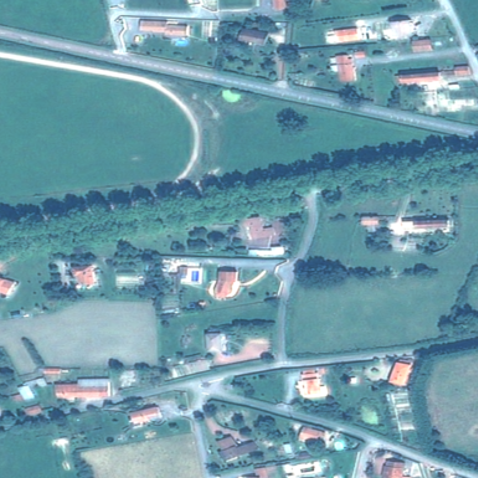}

\end{subfigure}
\begin{subfigure}[t]{0.65\linewidth}
\centering
Feature responses

\begin{subfigure}{0.49\linewidth}
\centering
\includegraphics[scale=0.17]{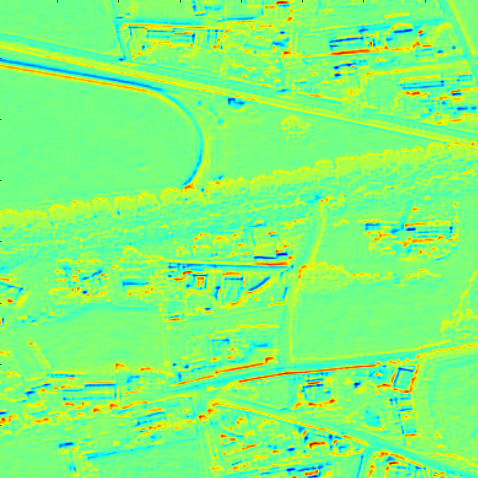}

{\small$N_1$}
\end{subfigure}
\begin{subfigure}{0.49\linewidth}
\centering
\includegraphics[scale=0.17]{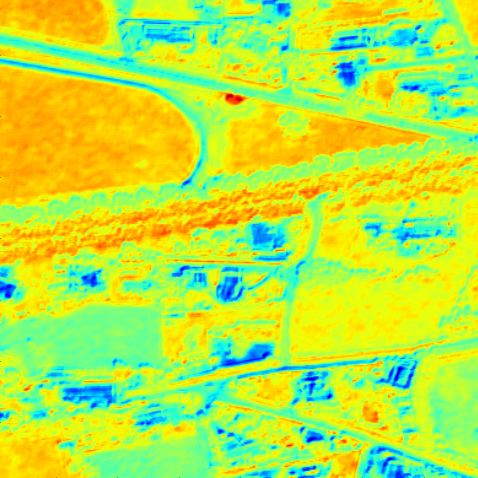}

{\small$N_2$}
\end{subfigure}
\end{subfigure}
\caption{$N_1$ acts like a gradient operator in the North direction and $N_2$ highlights green vegetation.}
\end{subfigure}
\caption{Feature responses (red: high, blue: low) to selected $M_i$ and $N_j$ filters, applied to the heat maps and input image respectively (see Eq.~\ref{eq:functionals_simplified}). }
\label{f:filters}
\end{figure}

 To interpret the meaning of the filters learned by the RNN we plot the map of responses of a sample input to the different filters. We here show some examples.
Fig.~\ref{f:filters}(a)  illustrates fragments of  heat maps of the \emph{building} and \emph{road} classes,  and the responses to two of the filters $M_i$ learned by the RNN. When analyzing these responses we can observe that they act as gradients in different directions, confirming the expected behavior.
Fig.~\ref{f:filters}(b) illustrates a fragment of the color image and its response to two filters $N_j$. One of them acts as a gradient operator an the other one highlights green vegetation, suggesting that this information is used to enhance the classification maps.

\section{Concluding remarks}

 In this work we presented an RNN that learns how to refine the coarse output of another neural network, in the context of pixelwise image labeling. The inputs are both the coarse classification maps to be corrected and the original color image. The output at every RNN iteration is an update to the classification map of the previous iteration, using the color image for guidance.

 
  Little human intervention is required, since the specifics of the refinement algorithm are not provided by the user but learned by the network itself. For this, we analyzed different iterative alternatives and devised a general formulation that can be interpreted as a stack of common neuron layers. At training time, the RNN discovers the relevant features to be taken both from the classification map and from the input image, as well as the function that combines them.
 
 The experiments on satellite imagery show that the classification maps are improved significantly, increasing the overlap of the foreground classes with the ground truth, and outperforming other approaches by a large margin. Thus, the proposed method not only detects but also outlines the objects.
To conclude, we demonstrated that RNNs succeed in learning iterative processes for classification enhancement tasks.

\section*{Acknowledgment}
All Pl\'eiades images are \copyright CNES (2012 and 2013), distribution \mbox{Airbus}~DS / SpotImage. The authors 
would like to 
thank 
the 
CNES for initializing and funding the study, and providing Pl\'eiades data.


\bibliographystyle{IEEEbib}
\bibliography{biblio}

\begin{IEEEbiography}[{\includegraphics[width=1in,
height=1.25in,clip,keepaspectratio]{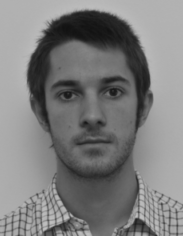}}]{Emmanuel Maggiori}
(S'15) received the Engineering degree in computer science from Central Buenos Aires Province National University (UNCPBA), Tandil, Argentina, in 2014. The same year he joined AYIN and STARS teams at Inria Sophia Antipolis-M\'editerran\'ee as a research intern in the field of remote sensing image processing. Since 2015, he has been working on his Ph.D.~within TITANE team, studying machine learning techniques for large-scale processing of satellite imagery. 
\end{IEEEbiography}

\begin{IEEEbiography}[{\includegraphics[width=1in,height=1.25in,clip,keepaspectratio]{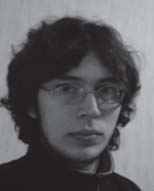}}]{Guillaume Charpiat}
 is a researcher at Inria Saclay (France) in the TAO team.
He studied Mathematics and Physics at the \'Ecole Normale Sup\'erieure (ENS Paris), and then Computer Vision and Machine Learning (at ENS Cachan), as well as Theoretical Physics.
His PhD thesis, in Computer Science, obtained in 2006, was on the topic of distance-based shape statistics for image segmentation with priors.
He then spent one year at the Max-Planck Institute for Biological Cybernetics (T\"ubingen, Germany), on the topics of medical imaging (MR-based PET prediction) and automatic image colorization. As a researcher at Inria Sophia-Antipolis (France), he worked mainly on image segmentation and optimization techniques. Now at Inria Saclay he focuses on Machine Learning, in particular on building a theoretical background for neural networks.
\end{IEEEbiography}

\begin{IEEEbiography}[{\includegraphics[width=1in,
height=1.25in,clip,keepaspectratio]{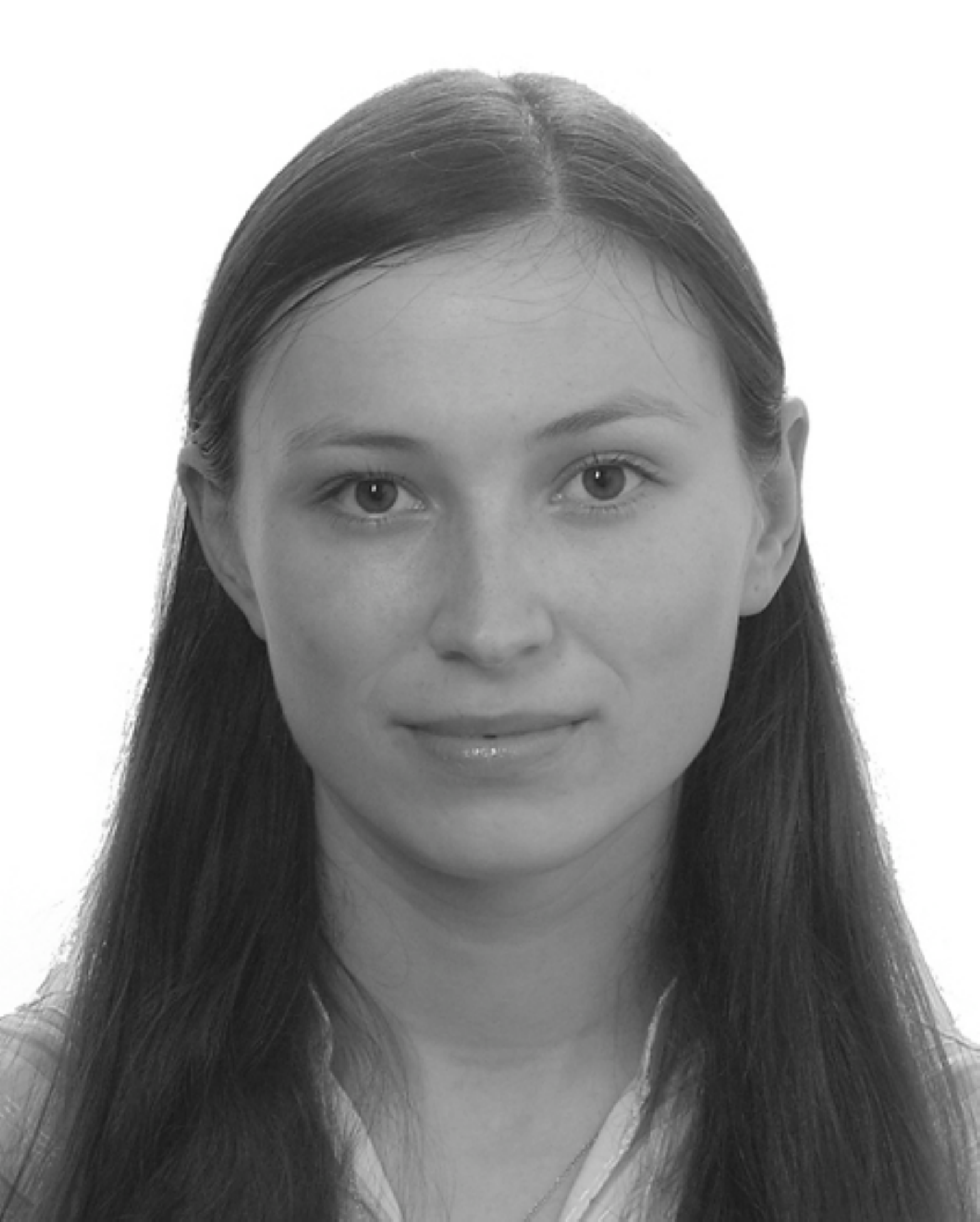}}]{Yuliya Tarabalka}
(S'08--M'10) received the B.S. degree in computer science from Ternopil Ivan Pul'uj State Technical University, Ukraine, in 2005 and the M.Sc. degree in signal and image processing from the Grenoble Institute of Technology (INPG), France, in 2007. She received a joint Ph.D. degree in signal and image processing from INPG and in electrical engineering from the University of Iceland, in 2010.

From July 2007 to January 2008, she was a researcher with the Norwegian Defence Research Establishment, Norway. From September 2010 to December 2011, she was a postdoctoral research fellow with the Computational and Information Sciences and Technology Office, NASA Goddard Space Flight Center, Greenbelt, MD. From January to August 2012 she was a postdoctoral research fellow with the French Space Agency (CNES) and Inria Sophia Antipolis-M\'editerran\'ee, France. She is currently a researcher with the TITANE team of Inria Sophia Antipolis-M\'editerran\'ee. Her research interests are in the areas of image processing, pattern recognition and development of efficient algorithms. She is Member of the IEEE Society.
\end{IEEEbiography}

\begin{IEEEbiography}[{\includegraphics[width=1in,height=1.25in,clip,keepaspectratio]{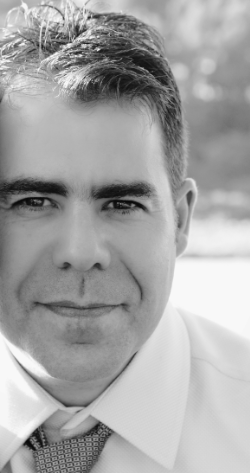}}]{Pierre Alliez}
Pierre Alliez is Senior Researcher and team leader at Inria Sophia-Antipolis - Mediterranee. He has authored scientific publications and several book chapters on mesh compression, surface reconstruction, mesh generation, surface remeshing and mesh parameterization. He is an associate editor of the Computational Geometry Algorithms Library (http://www.cgal.org) and an associate editor of the ACM Transactions on Graphics. He was awarded in 2005 the EUROGRAPHICS young researcher award for his contributions to computer graphics and geometry processing. He was co-chair of the Symposium on Geometry Processing in 2008, of Pacific Graphics in 2010 and Geometric Modeling and Processing 2014. He was awarded in 2011 a Starting Grant from the European Research Council on Robust Geometry Processing.
\end{IEEEbiography}

\end{document}